\documentclass[sigconf,10pt,nonacm]{acmart}

\usepackage{amsmath,amssymb,amsfonts}
\usepackage{mathtools}
\usepackage{algorithmic}
\usepackage{graphicx}
\usepackage{siunitx}
\usepackage{pifont}
\usepackage{textcomp}
\usepackage{xcolor}
\usepackage{fancyhdr}
\usepackage{chemformula}
\usepackage{graphicx}
\usepackage{setspace}
\usepackage{enumitem}
\usepackage{float}
\usepackage{url}
\usepackage{tabularx}
\usepackage{comment}
\usepackage{balance}
\usepackage{booktabs}
\usepackage{multirow}
\usepackage{titlesec}
\usepackage{enumitem}
\usepackage{titlecaps}
\usepackage{subcaption}
\usepackage{caption}
\usepackage{xcolor}
\usepackage{nicematrix}
\usepackage{array}
\usepackage{xspace}
\usepackage[utf8]{inputenc}
\usepackage{tcolorbox}
\usepackage{fancyhdr}
\usepackage[linesnumbered,ruled,vlined]{algorithm2e}

\renewcommand\footnotetextcopyrightpermission[1]{}

\author{Jiawei Hu}
\affiliation{%
  \institution{University of New South Wales}
  \institution{CSIRO}
  \country{Australia}}
\email{jiawei.hu@unsw.edu.au}

\author{Hong Jia}
\affiliation{%
  \institution{University of Melbourne}
  \country{Australia}}
\email{hong.jia@unimelb.edu.au}

\author{Mahbub Hassan}
\affiliation{%
  \institution{University of New South Wales}
  \country{Australia}}
\email{mahbub.hassan@unsw.edu.au}

\author{Lina Yao}
\affiliation{%
  \institution{CSIRO}
  \country{Australia}}
\email{lina.yao@data61.csiro.au}

\author{Brano Kusy}
\affiliation{%
  \institution{CSIRO}
  \country{Australia}}
\email{brano.kusy@data61.csiro.au}

\author{Wen Hu}
\affiliation{%
  \institution{University of New South Wales}
  \country{Australia}}
\email{wen.hu@unsw.edu.au}
\renewcommand{\shortauthors}{Hu, et al.}
\date{}

\begin{document}

\title{LightLLM: A Versatile Large Language Model for Predictive Light Sensing}
\renewcommand\abstractname{\textsc{ABSTRACT}}

\begin{abstract}
We propose LightLLM, a model that fine tunes pre-trained large language models (LLMs) for light-based sensing tasks. It integrates a sensor data encoder to extract key features, a contextual prompt to provide environmental information, and a fusion layer to combine these inputs into a unified representation. This combined input is then processed by the pre-trained LLM, which remains frozen while being fine-tuned through the addition of lightweight, trainable components, allowing the model to adapt to new tasks without altering its original parameters. This approach enables flexible adaptation of LLM to specialized light sensing tasks with minimal computational overhead and retraining effort. We have implemented LightLLM for three light sensing tasks: light-based localization, outdoor solar forecasting, and indoor solar estimation. Using real-world experimental datasets, we demonstrate that LightLLM significantly outperforms state-of-the-art methods, achieving 4.4x improvement in localization accuracy and 3.4x improvement in indoor solar estimation when tested in previously unseen environments. We further demonstrate that LightLLM outperforms ChatGPT-4 with direct prompting, highlighting the advantages of LightLLM's specialized architecture for sensor data fusion with textual prompts.
 
\end{abstract}
\maketitle


\section{INTRODUCTION}

\begin{figure}
    \centering
    \includegraphics[width=\linewidth]{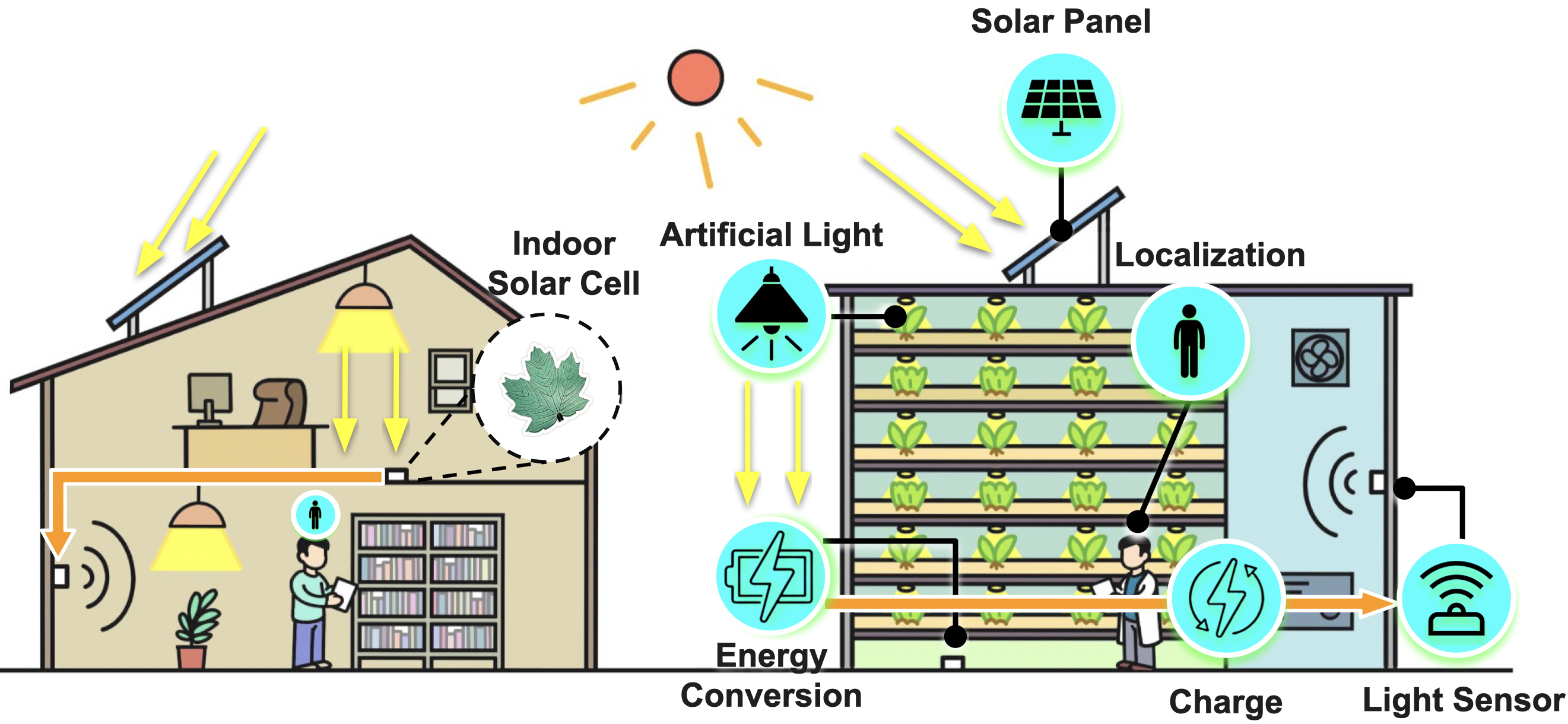}
    \vspace{-1em}
    \caption{Illustration of PLS tasks in the smart building system (left)~\cite{wang2017review}, and vertical farming system (right)~\cite{benke2017future}: they are equipped with outdoor solar panels to harness natural energy, self-sustaining sensors is deployed for localization task and powered by indoor decorative solar cells.}
    \label{fig:illu}
    \vspace{-1.5em}
\end{figure}

Predictive light sensing (PLS) is becoming increasingly significant as the demand for intelligent and sustainable systems grows. For example, by leveraging light sensor signatures, PLS can accurately infer the location of individuals in indoor spaces, enabling seamless {\bf indoor positioning} using light spectral information (LSI) where GPS signals are unreliable~\cite{hu2023iris} in smart buildings (see Figure~\ref{fig:illu} Left). Moreover, {\bf outdoor solar forecasting}~\cite{inman2013solar} for predicting future power generation of solar panels on rooftops based on weather conditions and sunlight availability and {\bf indoor solar estimation} for predicting the potential solar output at different locations within an indoor space~\cite{biswas2020solar}, such as a vertical farm (see Figure~\ref{fig:illu} Right)~\cite{benke2017future}, are becoming critical as buildings and smart devices shift towards sustainability.

Traditional approaches to solving PLS tasks rely on task-specific deep learning models tailored to each individual application~\cite{zhu2017deep}. These models require separate training, manual tuning, and frequent retraining to adapt to changing data distributions, leading to significant engineering overhead. Furthermore, they often struggle to generalize to unseen data or new environments~\cite{quinonero2022dataset}, which is particularly problematic in real-world scenarios where domain shifts due to changing conditions are common. Techniques like data augmentation~\cite{goodfellow2020generative} and transfer learning~\cite{zhuang2020comprehensive} are often employed to mitigate this issue. However, generating high-quality synthetic data is resource-intensive, and transfer learning is limited when the target task diverges significantly from the source domain. Consequently, these models remain both computationally expensive and inflexible, restricting their scalability across diverse PLS tasks.

Large Language Models (LLMs), with their vast pre-trained knowledge and cross-domain generalization capabilities, present new opportunities to address the challenges outlined above. Inspired by the success of models like ChatGPT~\cite{openai2023gpt4}, which have demonstrated exceptional performance in reasoning and answering complex questions across diverse contexts in natural language processing, LLMs show promise for offering a unified solution across various IoT applications, including light sensing tasks.

However, simply using LLMs with textual prompts, as explored in recent studies~\cite{ji2024hargpt,yang2024you}, is insufficient for these tasks. LLMs are not inherently designed to process sensor data, such as light spectral information or time-series solar output, nor can they generate non-text outputs required for specialized applications like location classification or energy prediction. While recent efforts have extended LLMs beyond traditional language tasks—such as NetLLM~\cite{wu2024netllm} for networking and MentalLLM~\cite{xu2024mental} for mental health assessment—these approaches still lack the ability to capture the domain-specific spatial, spectral, and environmental nuances critical for effective PLS sensing across different scenarios and environments.

We propose LightLLM, an innovative model designed to tackle light-based sensing tasks by leveraging the strengths of a pre-trained LLM. It integrates three key components: a task-specific sensor data encoder that extracts meaningful features from sensor inputs, a task-specific natural language prompt that provides contextual information about the environment, and a fusion layer that combines both sensor features and contextual prompts into a unified feature vector. This fused representation is then processed by the pre-trained LLM, which remains entirely frozen but fine-tuned using the LoRA (Low-Rank Adaptation) method. This unique architecture allows LightLLM to efficiently utilize the pre-existing general knowledge of LLMs while seamlessly integrating domain-specific sensor data, enabling it to handle specialized light sensing tasks with high adaptability and minimal retraining effort.

Key contributions of this paper can be summarized as follows:
\begin{itemize}
\setlist[itemize]{left=0pt}
    \item We present LightLLM, a general-purpose framework for adapting LLM to a wide range of light-based tasks with minimal effort. To our knowledge, this is the first attempt to use LLM for PLS.

    \item We conduct comprehensive real-world experiments to validate the effectiveness of LightLLM for three distinct PLS tasks, light-based indoor localization, outdoor solar forecasting, and indoor solar estimation. Our results show the superior performance of LightLLM in terms of accuracy and genralization compared to state-of-the-art methods.

    \item We provide an extensive ablation study to evaluate the impact of each component within LightLLM, offering insights into its adaptability and performance across different tasks.
\end{itemize}

\section{Methodology}
\label{sec:method}
\subsection{System Overview}

\begin{figure}
    \centering
    \includegraphics[width=\linewidth]{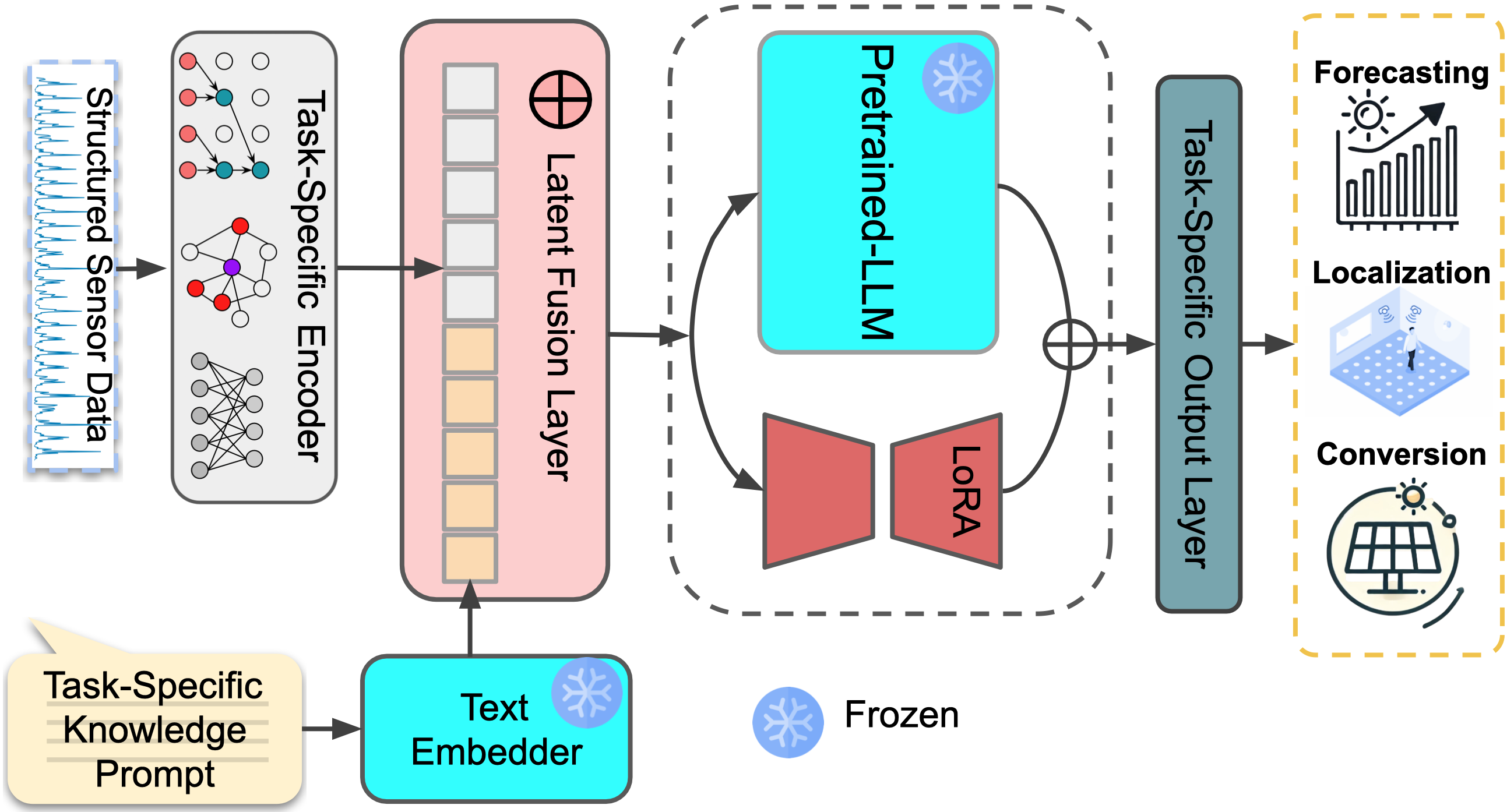}
    \vspace{-1em}
    \caption{Architecture of LightLLM.}
    \label{fig:overview}
    \vspace{-1em}
\end{figure}

Our proposed framework, \textbf{LightLLM}, integrates advanced encoders for various data modalities with the capabilities of LLMs to address the challenges of diverse tasks. The framework, as illustrated in Figure~\ref{fig:overview}, comprises five key components as followed.

\begin{itemize}[leftmargin=*, labelsep=0.5em]
    \item \textbf{Task-Specific Encoders}: Tailored to individual tasks, such as Graph Neural Networks (GNNs)~\cite{wu2020comprehensive} for localization, Temporal Convolutional Networks (TCNs)~\cite{lea2017temporal} for forecasting, and Convolutional Neural Networks (CNNs)~\cite{li2021survey} for energy estimation. For tasks like indoor localization, a custom Knowledge Graph (KG) is integrated to represent a richer context, enabling improved interpretation of sensor data.
    
    \item \textbf{Task-Specific Knowledge Prompts}: To provide structured contextual information, LightLLM employs task-specific prompts that encode domain knowledge essential to each task. These prompts are embedded as textual inputs, aligning sensor data properties and environmental factors with the LLM’s knowledge base.
    
    \item \textbf{Latent Fusion Layer (LFL)}: A specialized component that integrates the encoded data with latent representations from the LLM. Task-specific prompts are leveraged to enhance the knowledge representation of sensor data.
    
    \item \textbf{LoRA}: An efficient fine-tuning mechanism that adapts the LLM to specific tasks with minimal computational overhead.
    
    \item \textbf{Output Projection Layer}: The final layer that projects the combined representations into the desired output format, ensuring compatibility with task-specific requirements.
\end{itemize}

\subsection{Knowledge Graph for Enhanced Data Representation}
\label{sec:KG}

In LightLLM, the custom KG is designed to capture complex, spatially dependent relationships that are critical for tasks such as LSI indoor localization. By focusing on relevant sensor interactions, the KG enhances the interpretability and efficiency of GNN-based encoders by concentrating on task-specific correlations.

KGs have demonstrated their effectiveness in representing structured data across various domains. Recent research highlights their ability to improve task performance by organizing information into a structured, relational format~\cite{blagec2022curated}. In the context of smart building systems, standard ontologies like Brick~\cite{balaji2016brick} provide a schema for smart building metadata but fall short in addressing the specific needs of tasks such as LSI indoor localization. For example, such tasks require detailed spatial relationships informed by sensor orientation, light angles, and the presence of obstacles. To bridge this gap, we design a custom KG that captures these unique dependencies, enabling enhanced task performance.

The KG consists of nodes representing sensors and light sources, each with properties such as coordinates $(x, y, z)$, orientation, and detection range. The detailed \textit{Node Connection Algorithm} is outlined in Algorithm~\ref{alg:node_connection}, which constructs a KG by identifying meaningful connections between sensors and light sources based on spatial criteria. It consists of two phases as follows. In phase 1,
for each pair of sensors $(s_i, s_j)$ (Line 2), the Euclidean distance is computed (Line 3). If the distance is within $distance\_threshold$, their fields of view (FOVs) are calculated (Lines 5–6). A ``correlated'' edge is added if their FOVs overlap (Line 7) and no obstacles block the path (Line 8).
In phase 2, for each sensor $s_i$ and light source $l_k$ (Line 10), the horizontal distance ($d_h$) and vertical difference ($d_v$) are calculated (Lines 11–12). An edge is created if $d_h$ and $d_v$ are less than or equal to the thresholds (Line 13) and no obstacles exist (Line 14).

\begin{figure}
    \centering
    \includegraphics[width=0.8\linewidth]{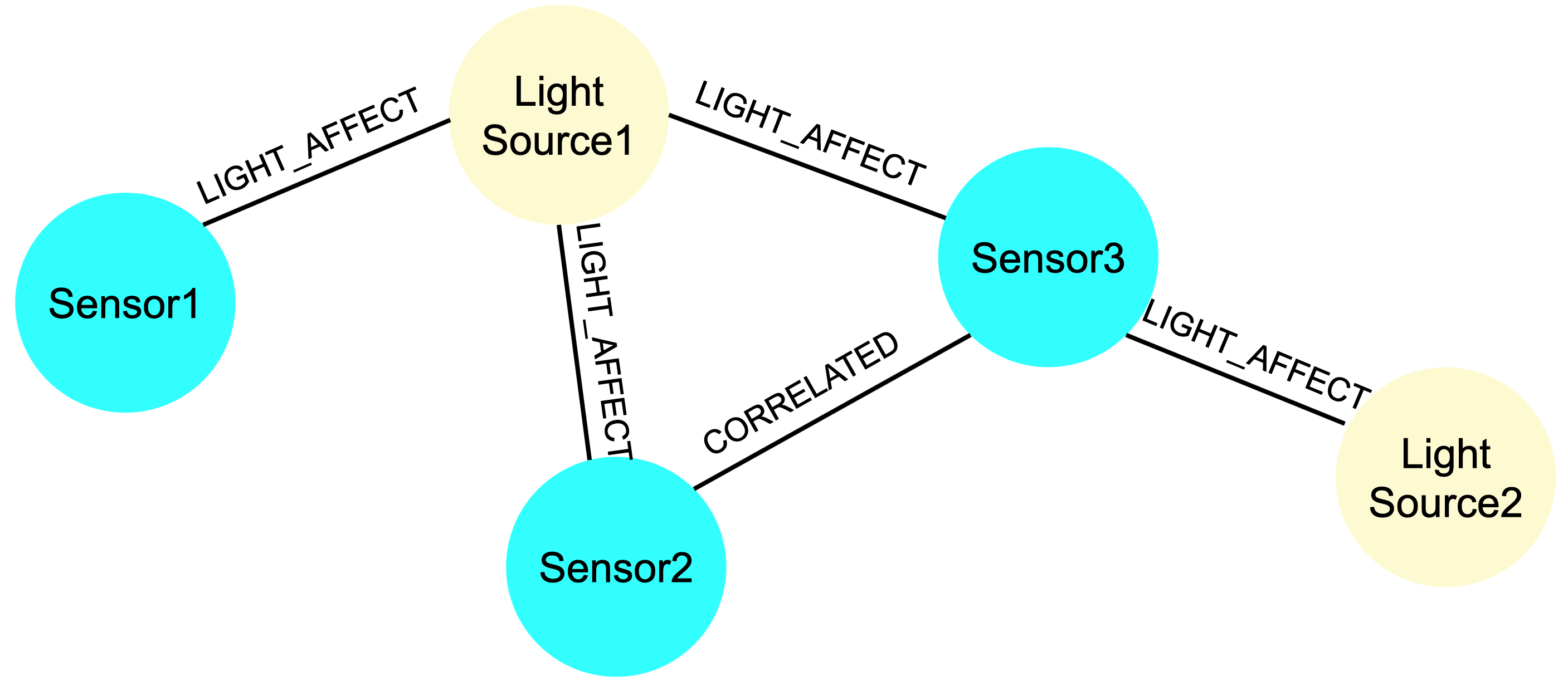}
    \vspace{-1em}
    \caption{Example of a KG illustrating sensor and light source interactions.}
    \label{fig:KG}
    \vspace{-1em}
\end{figure}

Figure~\ref{fig:KG} illustrates an example KG: Sensor 3 is influenced by both Light Sources 1 and 2, while Sensors 1 and 2 are influenced solely by Light source 1. Since Sensors 2 and 3 are spatially correlated, any change in illumination—such as movement by occupants—is likely to affect readings from both sensors. An obstacle positioned between Sensor 1 and Sensors 2/3 blocks any direct correlation between them.

Our KG includes obstacle-free connections only, and focuses on meaningful connections to reduce complexity. Furthermore, it enables
dynamic interactions based on FOV and spatial criteria to ensure realistic modeling.
This approach ensures the KG effectively captures relationships critical for enhancing LightLLM’s interpretation of sensor data.

\begin{algorithm}
\small
\caption{Node Connection Algorithm}\label{alg:node_connection}
\KwData{Set of sensors $S = \{s_1, s_2, ..., s_n\}$, set of light sources $L = \{l_1, l_2, ..., l_m\}$, set of obstacles $O$, field of view angle $angle$, distance threshold $distance\_threshold$, detection range $range$, vertical threshold $vertical\_threshold$}
\KwResult{Set of connections between sensors and light sources}

$connections \gets \{\}$ \;

\For{each pair of sensors $(s_i, s_j)$ where $i \neq j$}{
    $d \gets$ Distance($s_i$, $s_j$) \;
    
    \If{$d \leq distance\_threshold$}{
        $fov_i \gets$ CalculateFOV($s_i$, $angle$, $range$) \;
        $fov_j \gets$ CalculateFOV($s_j$, $angle$, $range$) \;

        \If{FOVOverlap($fov_i$, $fov_j$)}{
            \If{NoObstacle($s_i$, $s_j$, $O$)}{
                Append $(s_i, s_j, \texttt{CORRELATED})$ to $connections$ \;
            }
        }
    }
}

\For{each sensor $s_i$ and light source $l_k$}{
    $d_h \gets$ HorizontalDistance($s_i$, $l_k$) \;
    $d_v \gets$ VerticalDifference($s_i$, $l_k$) \;
    
    \If{$d_h \leq distance\_threshold$ and $d_v \leq vertical\_threshold$}{
        \If{NoObstacle($s_i$, $l_k$, $O$)}{
            Append $(s_i, l_k, \texttt{LIGHT\_AFFECTS})$ to $connections$ \;
        }
    }
}

\Return $connections$ \;
\end{algorithm}
\vspace{-1em}

\subsection{Task-Specific Encoders}

The task-specific encoders in LightLLM are designed to extract meaningful representations from diverse types of input data, ensuring that the most appropriate encoding techniques are applied. 

\subsubsection{Localization Encoder: Graph Neural Network}
For \textbf{indoor localization}, we employ a GNN to interpret spatial relationships among light sensors. The GNN leverages a graph structure where nodes represent sensor readings, and edges represent spatial relationships, as detailed in Section~\ref{sec:KG}. By aggregating information from neighboring nodes, the GNN captures inter-sensor dependencies, enhancing spatial awareness and improving localization accuracy.

\subsubsection{Forecasting Encoder: Temporal Convolutional Network}

For the task of \textbf{solar energy forecasting}, we employ a TCN to handle the time-series data from outdoor solar cells. TCNs are ideal for capturing long-range dependencies in time-series data due to their architecture, which combines causal convolutions and dilation.

\subsubsection{Energy Estimation Encoder: Convolutional Neural Network}

To estimate \textbf{indoor solar energy output}, we use a CNN to analyze spectral sensor data. The CNN extracts hierarchical features across different wavelengths, helping identify both low-level and high-level patterns in the spectral data. This capability is critical for modeling the complex interactions between light spectra and solar cell energy output.

\subsection{Task-Specific Knowledge Prompts}
For each task, a \textbf{task-specific prompt} is generated to provide additional contextual knowledge. These prompts embed domain-specific features such as spatial information for localization or environmental conditions for solar forecasting. This added context helps guide the model in understanding the relationship between input features.

As shown in Figure~\ref{fig:promptexmaple}, each task-specific prompt follows a structured template, which includes:

\begin{itemize}[leftmargin=*, labelsep=0.5em]
    \item Dataset Description: This section outlines the key characteristics of the dataset being used, such as the type of data (e.g., spectral wavelengths) and any relevant factors.
    \item Task Description: A brief explanation of the task.
    \item Data Organization: Details on how the data is structured, which helps the model understand the format of inputs.
    \item Key Input Characteristics: This includes important statistical properties or patterns in the data, such as temporal dependencies or spatial information, that are essential for the task.
\end{itemize}

It ensures that LightLLM leverages domain knowledge tailored to the unique challenges of each task. For example, in unseen environments where sensor layouts or lighting conditions might differ from the training set, the pre-trained LLM already possesses generalized knowledge about spatial relationships and environmental patterns. When combined with the specific spatial configurations provided in the prompt, LightLLM can better understand how a sensor's position or a light source's distribution influences the readings, helping the inference of model.

These prompts are then embedded using the LLM’s text embedding mechanism, and the resulting embedding objects are integrated into the Latent Fusion Layer.

\begin{figure}[ht]
    \centering
    \begin{tcolorbox}[colback=blue!3!white, colframe=blue!3!white, width=\linewidth, rounded corners, arc=5mm, boxsep=0.5mm]
    \small
        \textbf{<Dataset Description>:} The Power Output Measurements dataset contains data from a 30-kW rooftop photovoltaic array, with power output logged every 1 minute. The dataset captures the power generation patterns under varying sunlight and weather conditions.\\
        \textbf{<Task Description>:} Forecast the next \{\(T\)\} steps using the previous \{\(P\)\} steps of historical data, accounting for variations in sunlight intensity and environmental factors.\\
        \textbf{<Data Organization>:} \{\(PV\_output\)\}.\\
        \textbf{<Key Input Characteristics>:} 
        \begin{itemize}[leftmargin=*, labelsep=0.5em]
            \item Sunlight intensity over the past period, \{\(min\_values\)\} to \{\(max\_values\)\},
            \item Cloud cover and weather variability, impacting solar output trends,
            \item Temporal patterns such as day-night cycles,
            \item Geographic influence from location-specific sunlight exposure at \{\(latitude\)\}, \{\(longitude\)\}.
        \end{itemize}
        
        The focus is on modeling how geographic and environmental factors drive solar output variability for more accurate predictions.
    \end{tcolorbox}
    \vspace{-1em}
    \caption{Task-Specific Knowledge Prompt Example: Solar Energy Forecasting Prompt.}
    \label{fig:promptexmaple}
    \vspace{-1em}
\end{figure}

\subsection{Adaptive Latent Fusion Layer}

\begin{figure}[htb]
    \centering
    \includegraphics[width=0.6\linewidth]{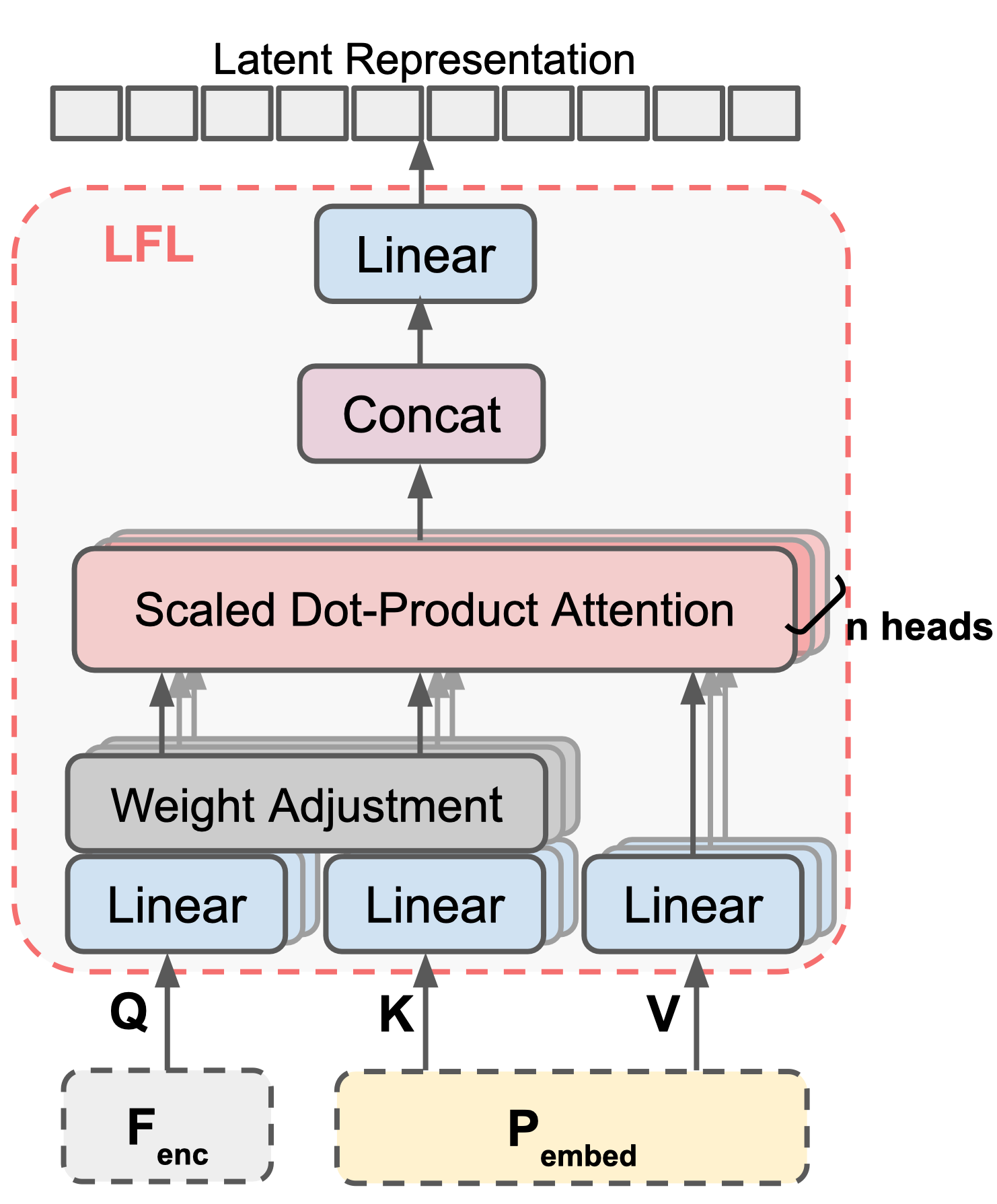}
    \vspace{-1em}
    \caption{Overview of Latent Fusion Layer.}
    \label{fig:LFL}
    \vspace{-1.5em}
\end{figure}

The LFL in LightLLM is a key component designed to integrate encoded features from task-specific encoders with prompt embeddings generated from task-specific knowledge prompts. This integration is achieved through an multi-head attention mechanism~\cite{vaswani2017attention}, allowing the model to leverage the strengths of both the feature encoders and the LLM, the structure is shown in Figure~\ref{fig:LFL}. The LFL ensures that the encoded data is aligned with the LLM's latent space, facilitating effective fusion of multimodal information.

The encoded features \( F_{\text{enc}} \) from the task-specific encoder and the embedded prompt \( P_{\text{embed}} \) are fused using a multi-head attention mechanism within the LFL. Specifically, the query \( Q \), key \( K \), and value \( V \) matrices are computed from the encoded features and the prompt embeddings as follows:

\[
Q = W_Q F_{\text{enc}}, \quad K = W_K P_{\text{embed}}, \quad V = W_V P_{\text{embed}}.
\]

Here, the prompt embeddings \(P_{\text{embed}}\) encapsulate structured, task-specific knowledge that complements the encoded sensor features. By assigning \(P_{\text{embed}}\) as both the key and value, the attention mechanism identifies relevant knowledge and integrates it into the encoded features \(F_{\text{enc}}\). The key (\(K\)) assesses relevance, while the value (\(V\)) provides the information to enhance the final representation.

Each head in the multi-head attention mechanism works with different learned weight matrices \(W_Q\),\(W_K\) and \(W_V\). It splits the matrices into \(n_{heads}\) parallel subspace, computes attention separately for each head, and then concatenates the results.

In addition to the standard attention mechanism, the LFL incorporates a weight adjustment mechanism to control the contribution of the encoded data and the text embeddings. This is achieved through learnable parameters \( \alpha \) and \( \beta \), which adjust the weights of the target and source embeddings, respectively. The target embedding \( T \) and source embedding \( S \) are projected and scaled as follows:

\begin{equation}
T' = \alpha \cdot Q, \quad S' = \beta \cdot K,
\end{equation}


where \( T' \) and \( S' \) are the weighted target and source embeddings. The attention scores are then computed using the scaled dot-product of the weighted embeddings:

\begin{equation}
\text{Scores} = \text{softmax}\left(\frac{T' {S'}^T}{\sqrt{d_k}}\right),
\end{equation}

where \( d_k \) is the dimension of the key vectors. Once the fusion process is completed, the resulting representation is passed into the LLM to generate the reprogrammed latent representation. The resulting latent representation combines the strengths of the encoded sensor data and the prompt knowledge, providing a task-specific feature that is well-suited for downstream tasks in PLS.

The LFL also includes normalization and dropout layers to ensure stable training and prevent overfitting. LFL effectively leverages both pre-trained knowledge and task-specific information, allowing LightLLM to produce highly accurate, adaptable outputs across various PLS applications.

\subsection{Low-Rank Adaptation}

To efficiently adapt the pre-trained language model for our specific downstream tasks, we utilize LoRA~\cite{hu2021lora}, which introduces low-rank matrices around the original pre-trained LLM, allowing for dimensionality reduction and expansion, effectively performing a rank adaptation. During training, the parameters of the LLM remain frozen, while only the low-rank matrices \( A \) and \( B \) are trained. This significantly reduces the number of trainable parameters while preserving the model's capacity to adapt to new tasks.

Initially, the reduction matrix \( A \) is randomly initialized using a Gaussian distribution, while the expansion matrix \( B \) is initialized as a zero matrix. This setup ensures that the pre-trained behavior of the LLM is maintained at the beginning of training, with the adaptation gradually introduced as the matrices are trained.

The modified weight matrix \( W' \) is computed as follows:

\begin{equation}
W' = W + \alpha AB,
\end{equation}

where \( W \) is the pre-trained weight matrix (e.g., a frozen LLM). The dimensions of low-rank matrices \( A \in \mathbb{R}^{d \times r} \) and \( B \in \mathbb{R}^{r \times d} \) are significantly smaller than the full weight matrix \( W \in \mathbb{R}^{d \times d} \), significantly reducing the number of parameters being fine-tuned. 

During the forward pass, the input \( x \) is computed as follows:

\begin{equation}
W' x = W x + \alpha (AB)x.
\end{equation}

This approach is analogous to a residual connection, where the low-rank updates simulate the full fine-tuning process by introducing an adaptable layer on top of the pre-trained LLM. When the rank \( r \) equals the dimension of \( W \), LoRA effectively performs full fine-tuning, but with \( r \ll d \), LoRA provides a more computationally efficient way to adapt the model. In the inference phase, LoRA introduces minimal additional latency, with parallel computations for \( W x \) and \( (AB)x \).

In our model, we emperically chose the settings \( r = 64 \), \( \alpha = 64 \), and dropout of \( 0.1 \), to achieve an optimal balance between task-specific adaptation and preserving the pre-trained knowledge of the LLM, while maintaining computational efficiency.

\subsection{Task-Specific Output Layer}

The output layer of LightLLM is designed to handle the specific requirements of different tasks by employing task-specific heads. This ensures that the final model outputs are appropriate for each application.

For example, for the localization task, the output layer is configured as a probability classification head~\cite{goodfellow2016deep}. This head outputs a probability distribution over possible locations, allowing the model to predict the most likely position based on the input data. The classification head uses a softmax activation function to convert the model's logits into probabilities, thereby facilitating accurate location predictions.

The flexible design of this task-specific head allows LightLLM to be easily adapted to different tasks. By incorporating the appropriate headers, the model is able to provide accurate, task-relevant output, whether it involves classification, regression, or other output formats. This module ensures that the model's predictions are not only accurate but also customized to the unique characteristics of each PLS task.

\section{EVALUATION}
\label{sec:evaluation}

\subsection{Light Spectral Localization}
Indoor localization using light spectral information has proven to be a highly effective technique for achieving passive Visible Light Positioning (VLP) in indoor environments. This method leverages the unique properties of light in different spaces to estimate the location of users. One of the leading systems for spectral-based localization is $Iris$~\cite{hu2023iris}, which employs a conditional Generative Adversarial Network (cGAN) for data augmentation to enhance its performance. 

To evaluate the performance of LightLLM in indoor localization, we collected data from two distinct indoor environments: a large office and a apartment. In the office environment, we deployed 27 spectral sensors placed strategically throughout the 108 square meter area. In the apartment environment, we deployed 17 spectral sensors across the 25 square meter space. Each environment was carefully configured to capture spectral data over multiple sessions.

\begin{figure}[t]
    \centering
        \begin{subfigure}[b]{0.49\linewidth}
        \centering
        \includegraphics[width=\textwidth]{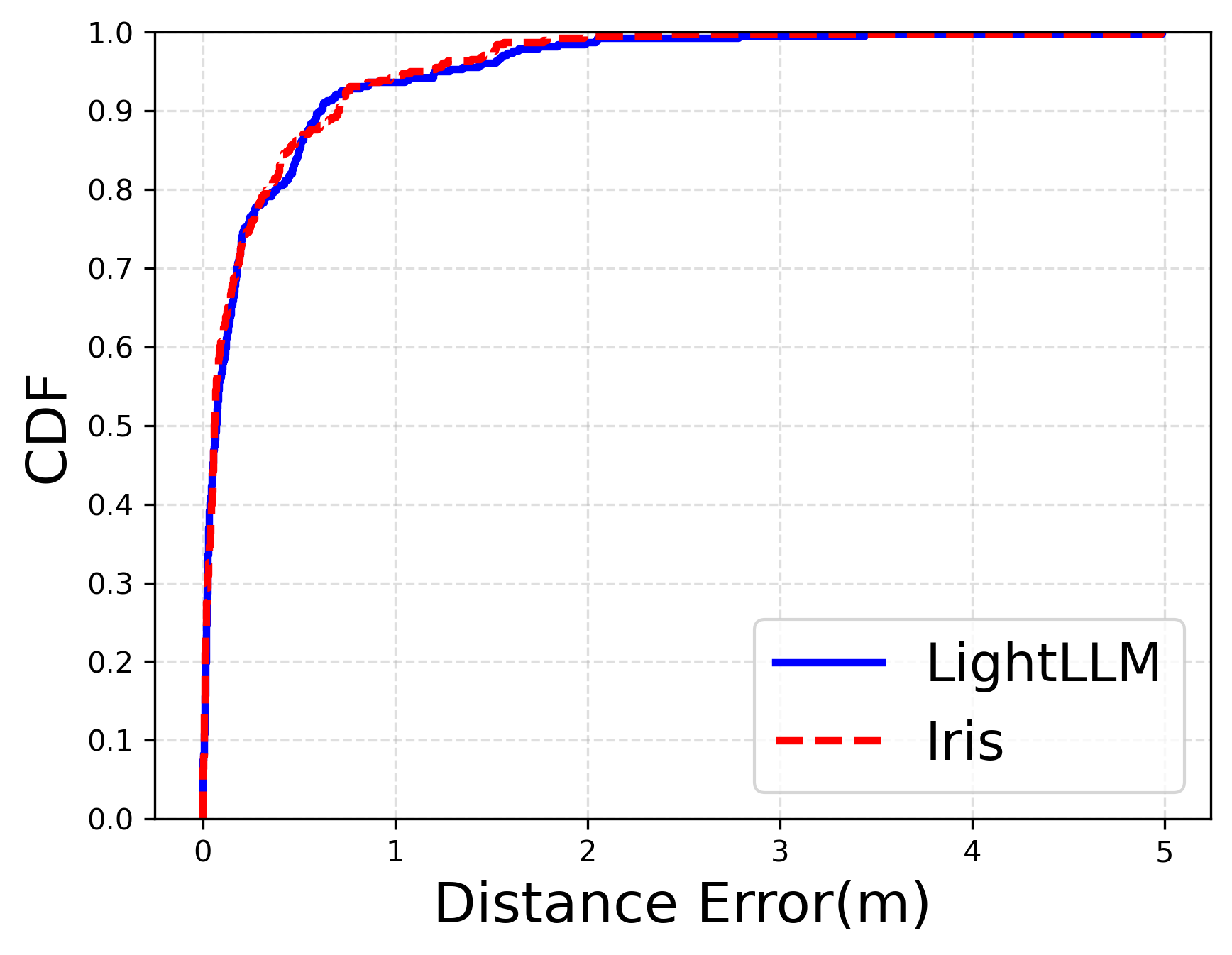}
        \caption{Apartment}
    \end{subfigure}
    \hspace{-0.4em}
    \begin{subfigure}[b]{0.49\linewidth}
        \centering
        \includegraphics[width=\textwidth]{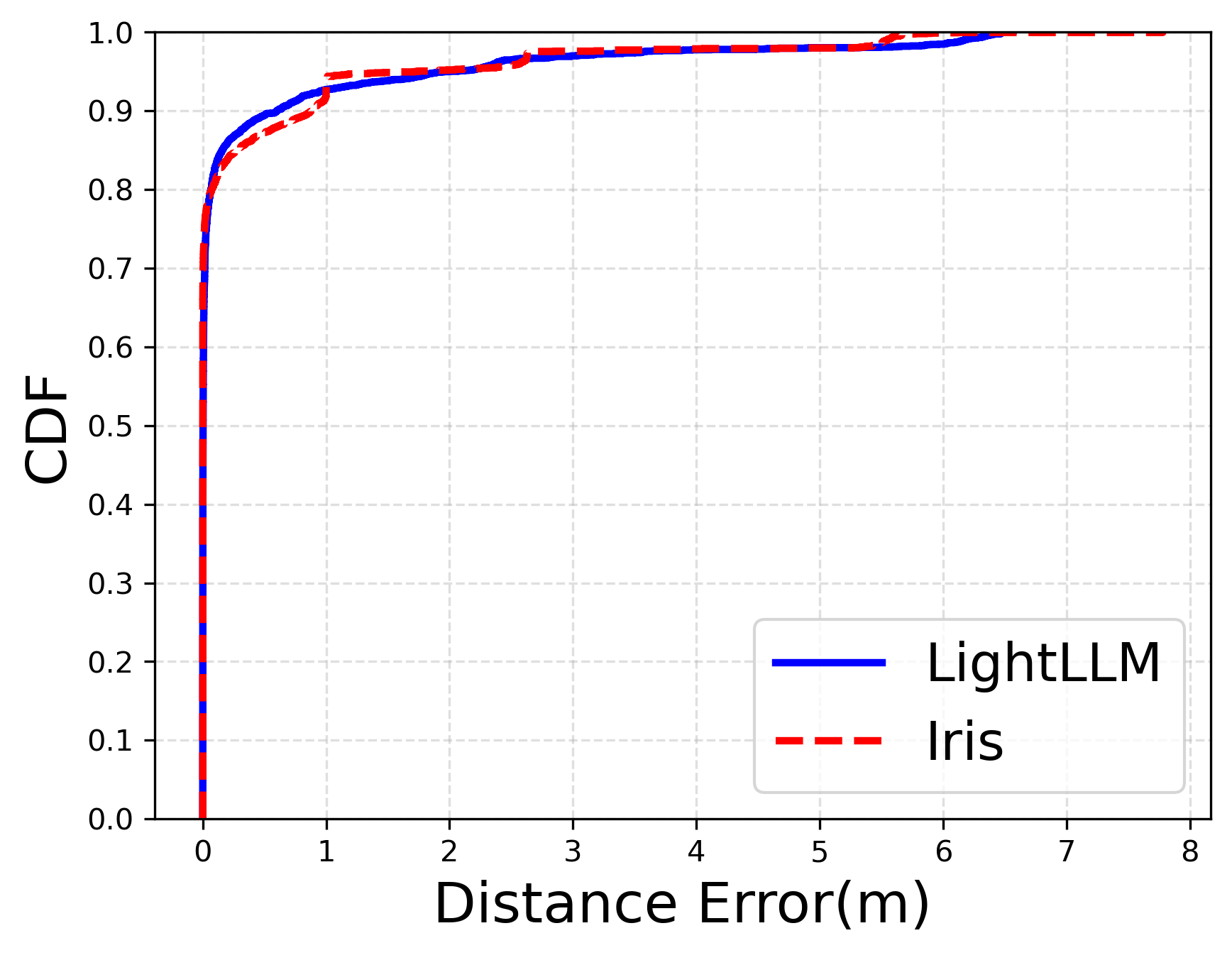}
        \caption{Office}
    \end{subfigure}
    \vspace{-1em}
    \caption{Localization error under \textit{seen} scenarios.}
    \label{fig:localizationseen}
    \vspace{-1em}
\end{figure}

In the \textbf{seen environment} tests, where both training and testing data are collected from the same indoor room, LightLLM shows comparable or slightly better performance than $Iris$ (see Figure~\ref{fig:localizationseen}). For example, in the 90th percentile error, where LightLLM achieves 0.608m vs 0.705m for $Iris$ in the apartment environment, and LightLLM achieves 0.89m, while Iris performs slightly worse at 0.60m of 90th percentile error in the office environment.

In the \textbf{unseen environment}, where the training data does not include the environment used in the test set (e.g., train with apartment's dataset and test with office's dataset). This setting could evalute how well the system generalizes to new environments, which is a significant challenge for traditional models like $Iris$.

\begin{figure}[t]
    \centering
        \begin{subfigure}[b]{0.49\linewidth}
        \centering
        \includegraphics[width=\textwidth]{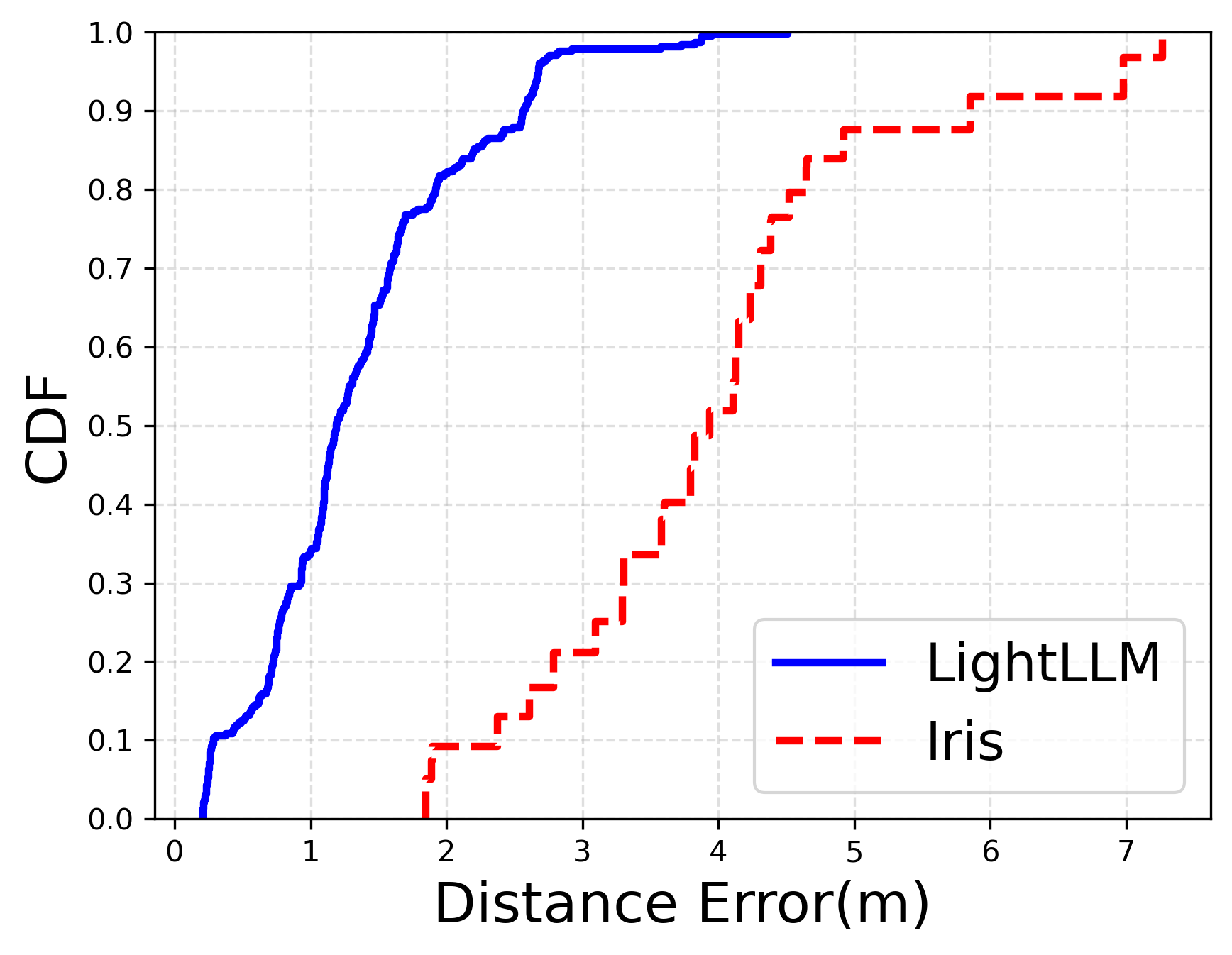}
        \caption{Apartment}
    \end{subfigure}
    \hspace{-0.4em}
    \begin{subfigure}[b]{0.49\linewidth}
        \centering
        \includegraphics[width=\textwidth]{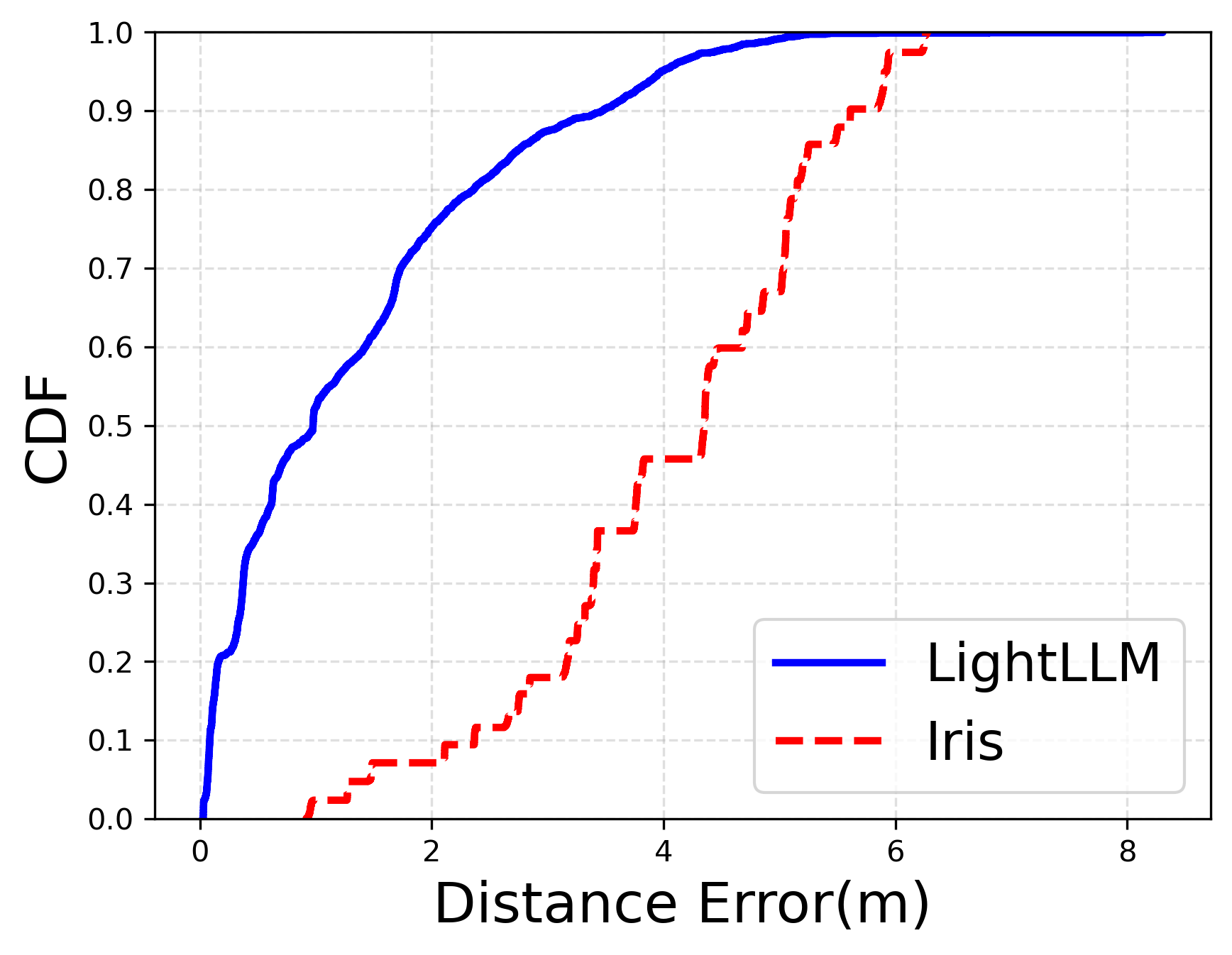}
        \caption{Office}
    \end{subfigure}
    \vspace{-1em}
    \caption{Localization error under \textit{unseen} scenarios.}
    \label{fig:localizationunseen}
    \vspace{-2em}
\end{figure}

The results in Figure~\ref{fig:localizationunseen} show that $Iris$ sees a dramatic increase in error in both environments, with a median error of 3.93m and 4.35m, respectively, a significant degradation from its seen performance. In contrast, LightLLM maintains far better generalization with a median error of only 0.98m in office and 1.19m in apartment, achieving 4.4x improvement. Similarly, 75th percentile and 90th percentile error further emphasize this difference, for example, LightLLM achieves 1.99m and 3.46m compared to 5.05m and 5.61m for $Iris$ in the office. This demonstrates LightLLM's superior ability to handle difficult and unseen localization tasks, where traditional method failed to work.

In summary, LightLLM exhibits competitive performance in seen environments but shows superior generalization capabilities in unseen environments. The ability of LightLLM to adapt to new environments with minimal degradation in performance highlights its advantage over traditional systems like $Iris$, which almost impossible to work in a completely new environment with unfamiliar data. This makes LightLLM a more robust and scalable solution for light spectral indoor localization, especially in real-world applications where environments are diverse and complex.

\subsection{Outdoor Solar Energy Forecasting}
Solar forecasting plays a crucial role in optimizing energy management systems and improving the reliability of renewable energy sources. Accurate forecasts of photovoltaic (PV) generation help ensure energy grid stability and enable more efficient integration of solar energy into the energy mix. To achieve these predictions, solar forecasting models typically rely on a variety of data models, including historical PV generation and environmental data, such as images of the sky that capture cloud movement.

The dataset used in this evaluation, SKIPP'D~\cite{nie2023skipp} includes over three years of PV power generation data. It covers a diverse range of weather conditions, providing an excellent opportunity to assess the performance of solar forecasting models across various scenarios. The dataset includes PV output data recorded at high temporal resolution, offering rich insights into power generation patterns over time.

To evaluate the performance of LightLLM, we employ several commonly used metrics in solar forecasting~\cite{gneiting2014probabilistic}:
\begin{itemize}[leftmargin=*, labelsep=0.5em]
\item \textbf{Continuous Ranked Probability Score (CRPS)}: A lower CRPS indicates more accurate probabilistic predictions.

\item \textbf{Forecast Skill (FS)}: This metric reflects the percentage improvement over a baseline method (e.g., smart persistence~\cite{da2021fundamentals}), with higher FS values representing better improvements.

\item \textbf{Winkler Score (WS)}: This metric evaluates the width of the prediction intervals at a given probability level, typically aiming for narrow intervals that still capture the true value. The lower WS indicates a better probabilistic forecast.
\end{itemize}
These metrics enable a thorough assessment of how well different forecasting models predict solar energy output under various conditions.

\begin{table*}
\centering
\caption{LightLLM vs. state-of-the-art for outdoor solar forecasting, with Smart Persistence as the baseline.}
\vspace{-1em}
\label{tab:forecast}
\begin{tabular}{c|ccc|ccc}
    \toprule
    \multirow{2}{*}{\textbf{Method}} & \multicolumn{3}{c|}{\textbf{Seen}} & \multicolumn{3}{c}{\textbf{Unseen}} \\ \cmidrule(lr){2-4} \cmidrule(lr){5-7}
                                     &   CRPS[kW]$\downarrow$ & FS[\%]$\uparrow$  &  WS$\downarrow$ & CRPS[kW]$\downarrow$ & FS[\%]$\uparrow$ & WS$\downarrow$ \\
    \midrule
    Smart Persistence            & 2.89 & 0 & - & 3.67 & 0 & - \\
    \midrule
    ConvLSTM      & 2.70 & 6.6 & 41.12 & 3.09 & 15.8 & 43.82 \\
    SkyGPT      & 2.49 & 13.8 & 26.72 & 2.81 & 23.4 & 26.70 \\
    TimeLLM      & 2.27 & 21.6 & 21.30 & 2.67 & 27.3 & 24.30 \\
    \textbf{LightLLM}      & \textbf{1.92} & \textbf{33.7} & \textbf{20.80} & \textbf{2.52} & \textbf{31.4} & \textbf{22.56} \\
    \bottomrule
\end{tabular}
\vspace{-1em}
\end{table*}


We compared LightLLM against several state-of-the-art methods, including traditional method ConvLSTM~\cite{shi2015convolutional}, and SkyGPT~\cite{nie2024skygpt}, a leading model that employs sky images in combination with U-Net for solar energy forecasting, and TimeLLM~\cite{jin2023time}, which also utilizes LLM for time-series predictions. All of them are comparing with a commonly used baseline method, smart persistence model~\cite{da2021fundamentals}. The results in the seen environment are shown in the Table~\ref{tab:forecast}. LightLLM outperforms SkyGPT in several key metrics, obtaining a CRPS of 1.92 kW, representing a 33.7\% improvement in FS, which is significantly higher than SkyGPT's performance. Furthermore, LightLLM records a WS of 20.80, demonstrating superior accuracy in matching the predicted solar output distribution to the actual data. Additionally, while TimeLLM offers solid performance, LightLLM consistently delivers better results across all metrics, highlighting the advantage of integrating task-specific encoders and the LFL in LightLLM.


\begin{figure}
    \centering
    \includegraphics[width=0.95\linewidth]{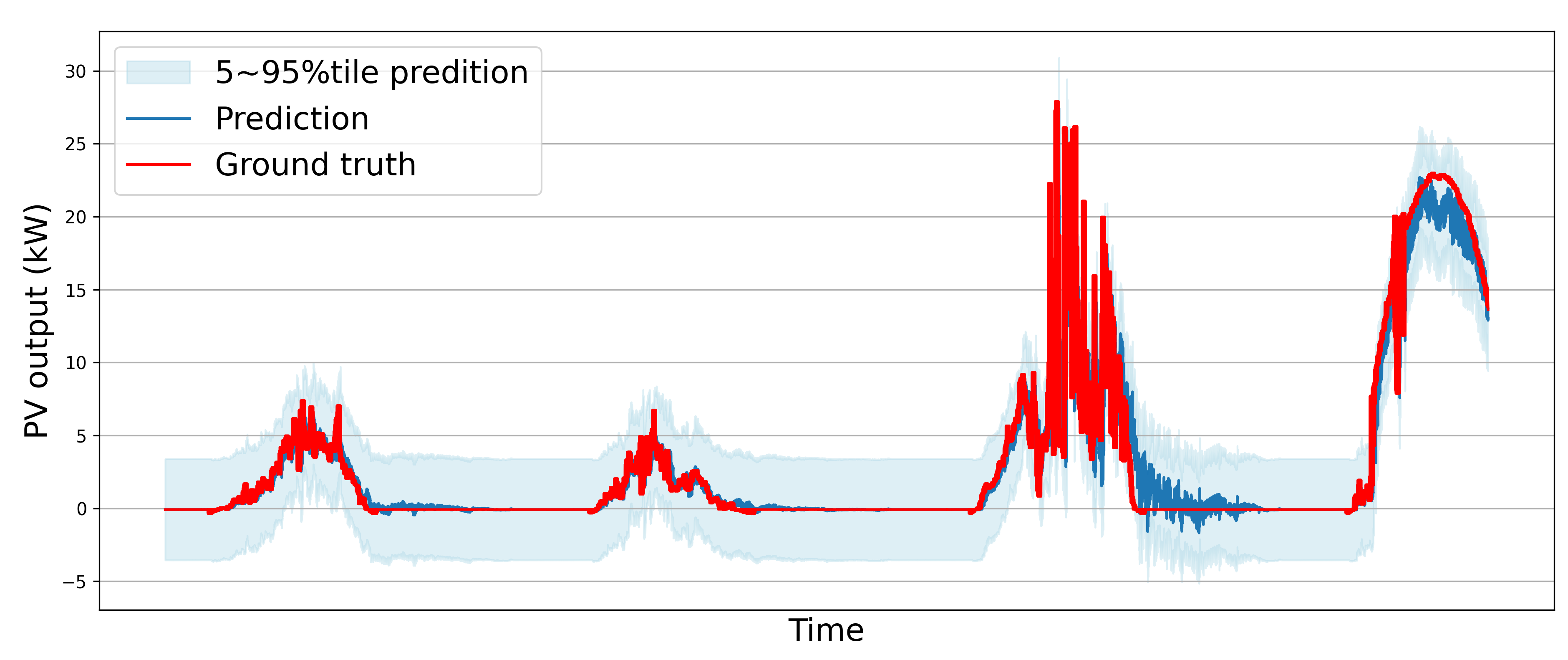}
    \vspace{-1em}
    \caption{Prediction curve of \textbf{LightLLM} and ground truth with prediction intervals (5-95\%) under unseen scenarios.}
    \label{fig:predvsgt}
    \vspace{-1em}
\end{figure}

In the unseen scenario, LightLLM is evaluated on a completely separate time period of data that the model has not encountered during training, specifically, we trained by using the data from 2017 Mar to 2019 Oct and tested by using data from 2019 Nov to 2019 Dec. It is intended to simulate real-world conditions to evaluate the final performance of the model.
By looking at the results shown in Table~\ref{tab:forecast} and Figure~\ref{fig:predvsgt}, LightLLM maintains its edge with a CRPS of 2.52 kW and a FS improvement of 31.4\%, outperforming SkyGPT (CRPS: 2.81 kW, FS: 23.4\%) and TimeLLM (CRPS: 2.67 kW, FS: 27.3\%). Furthermore, LightLLM records a WS of 22.56, again lower than SkyGPT’s 26.70 and TimeLLM’s 24.30, demonstrating its robustness. This indicates that LightLLM not only excels in seen conditions but also generalizes well in unseen settings.

The comparison clearly shows that LightLLM offers better performance across both seen and unseen environments, providing superior accuracy and generalization capabilities compared to other leading models such as SkyGPT and TimeLLM.

\subsection{Indoor Solar Energy Estimation}

\subsubsection{Challenges in Real-World Solar Energy Estimation. }
With the increasing deployment of solar cells for indoor energy harvesting applications, such as those used in low-power IoT devices~\cite{samsungTV} or used as interactive surfaces~\cite{meena2020pv}, presents an opportunity to harness light more effectively for energy generation in indoor spaces. Given that different types of solar cells exhibit unique responses to varying light conditions, optimizing their placement and understanding their energy output in real-world environments becomes critical. This optimization can greatly enhance the efficiency of solar energy systems in indoor settings.

Fortunately, existing technology allows us to simulate and collect spectral sensor data~\cite{hu2024lidarspectra}. However, the challenge lies in accurately translating this spectral data into solar energy output for each unique solar cell. While mathematical models exist to predict this conversion under idealized conditions, their accuracy diminishes in complex, real-world scenarios due to various discrepancies between theory and practice.

In theory, the short-circuit current density $J_{sc}$ is a parameter in assessing the performance of a solar cell. As described in~\cite{wright2012organic}, this value depends on the external quantum efficiency ($EQE$), which is the proportion of photo-generated electrons to incident photons, and the intensity of photons at each wavelength $P_{in}(\lambda)$. Following the approach in~\cite{ma2019solargest}, the total current generated by a solar cell can be expressed as:
\begin{equation}
\label{eq:solarcell}
\text{Current} = A \cdot \int_{\lambda_{min}}^{\lambda_{max}} a(\lambda) \cdot I(\lambda) \cdot \lambda \cdot d\lambda,
\end{equation}
given the spectral absorption rate \(a(\lambda)\), the incident spectrum \(I(\lambda)\), and the surface area \(A\) of the solar cell.

\begin{figure}
    \centering
    \includegraphics[width=0.8\linewidth]{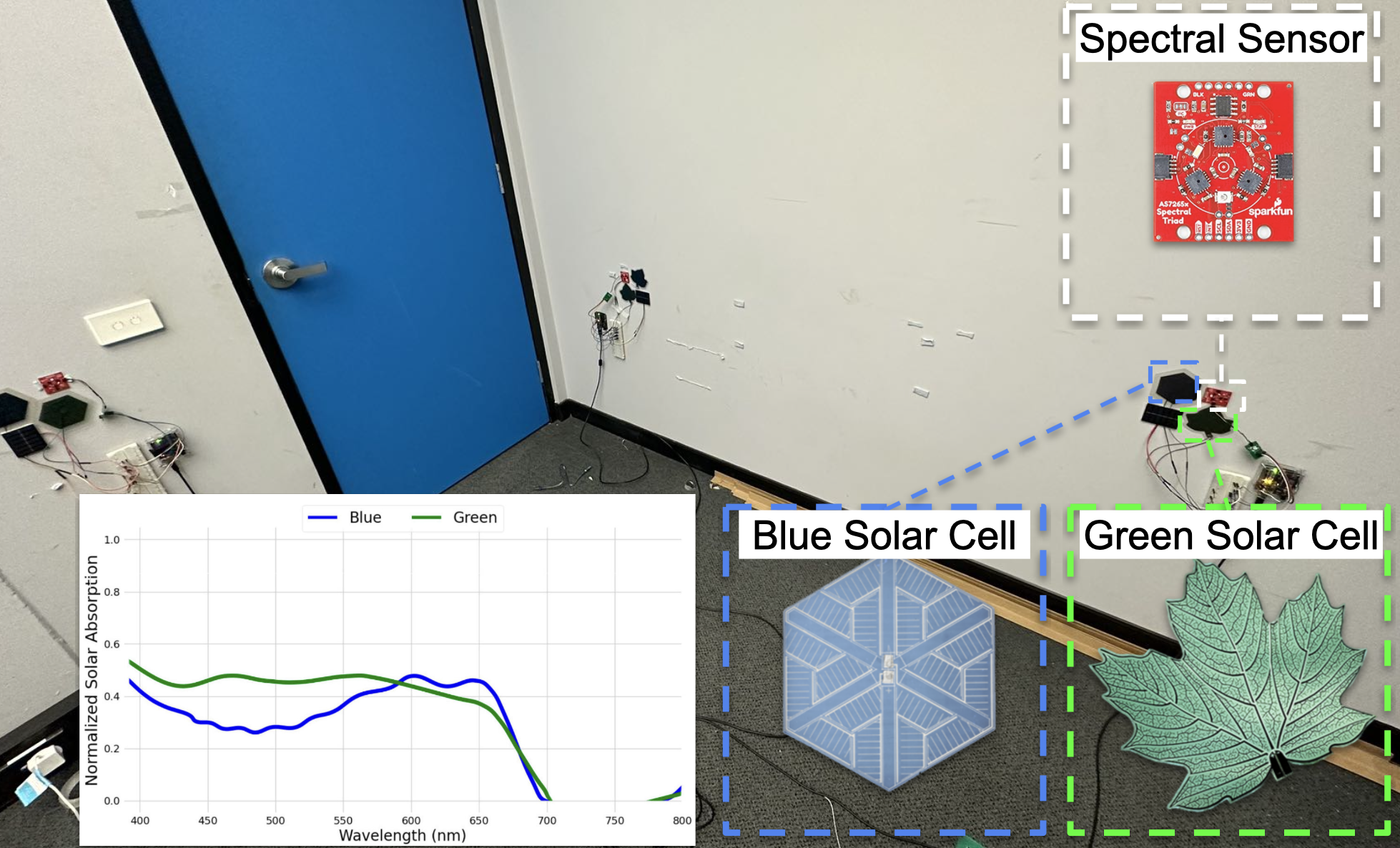}
    \vspace{-1em}
    \caption{Environment for deploying spectral sensors and semi-transparent solar cells with different absorption rates.}
    \label{fig:solarsetting}
    \vspace{-1em}
\end{figure}

\begin{table}[ht]
\centering
\caption{Actual vs. predicted photocurrent (mA) for solar cells at various indoor locations.}
\vspace{-1em}
\label{tab:SolarCellPhotocurrent}
\begin{tabular}{cccccc}
    \toprule
    \multirow{2}{*}{\textbf{Location}} & \multicolumn{2}{c}{\textbf{Blue Solar Cell}} & \multicolumn{2}{c}{\textbf{Green Solar Cell}} \\ 
    & \textbf{Actual} & \textbf{Predicted} & \textbf{Actual} & \textbf{Predicted} \\
    \midrule
    1 & 0.596 & 0.335 & 0.459 & 0.249 \\
    \midrule
    2 & 0.547 & 0.306 & 0.523 & 0.242 \\
    \midrule
    3 & 0.097 & 0.141 & 0.348 & 0.517 \\
    \bottomrule
    \textbf{Avg MAPE[\%]} &\multicolumn{2}{c}{44.31} & \multicolumn{2}{c}{49.34} \\
    \bottomrule
\end{tabular}
\vspace{-1em}
\end{table}

However, our real-world experiments revealed significant discrepancies between theoretical predictions and actual solar cell performance. In these tests, we deployed both spectral sensors (AS7265x) with 18 wavelength channels and solar cells with different absorption rates in an indoor room setup, as shown in Figure~\ref{fig:solarsetting}. Each solar cell's and spectral sensor's output was measured using the same Arduino device and synchronized to ensure consistency. We compared the measured outputs with predicted values from Eq.~(\ref{eq:solarcell}) and used Mean Absolute Percentage Error (MAPE) and Mean Squared Error (MSE) as the evaluation metrics, calculated as follows:
\begin{equation}
\text{MAPE} = \frac{1}{n} \sum_{i=1}^n \left| \frac{\text{Actual}_i - \text{Predicted}_i}{\text{Actual}_i} \right| \times 100,
\end{equation}
\begin{equation}
\text{MSE} = \frac{1}{n} \sum_{i=1}^n \left( \text{Actual}_i - \text{Predicted}_i \right)^2,
\end{equation}
where $n$ represents the total number of data samples, with \( n = 1,035 \) in our experiments.

From the results of Table~\ref{tab:SolarCellPhotocurrent}, we observed an average MAPE of \textbf{46.83\%} across different room configurations, while the individual MAPE values for each solar cell are as follows: Solar Cell Blue \textbf{44.31\%} and Solar Cell Green \textbf{49.34\%}. This high error rate underscores the limitations of the equation-based approach. 

\begin{figure}
    \centering
    \includegraphics[width=0.9\linewidth]{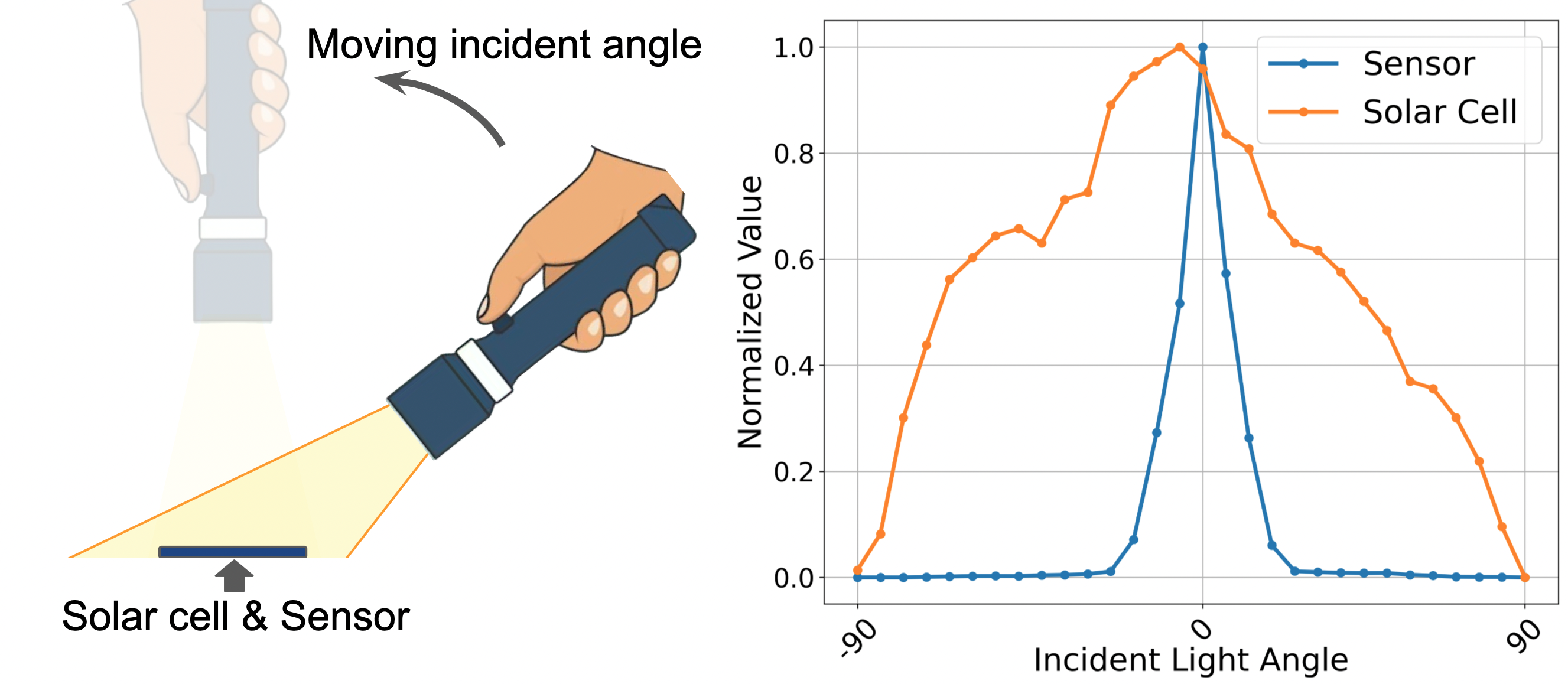}
    \vspace{-1em}
    \caption{Experiment results show the difference between the solar cell and light sensor readings when the light hits them from different angles.}
    \label{fig:solarsensordiffill}
    \vspace{-1em}
\end{figure}

One of the fundamental reasons we found during our experiments was the discrepancy between the behavior of spectral sensors and solar cells. Spectral sensors~\cite{as7265x} are designed with a defined FOV. In contrast, solar cells are not optimized for measuring light at specific angles but aim to maximize overall absorption across their surface area~\cite{parretta1998effects}. Solar cells perform most efficiently when light is incident perpendicular to the surface. However, as the angle of incidence deviates from the vertical axis, the efficiency of light absorption may decrease. To investigate this, we conducted experiments where both spectral sensors and solar cells were deployed at the same location and under different angles of light incidence, the result is shown in Figure ~\ref{fig:solarsensordiffill}.

Additionally, we observed several other factors that contributed to inaccuracies when applying the mathematical models to real-world scenarios. These factors include variations in the packaging of solar cells, which can affect the way light interacts with the cell surface, as well as manufacturing inconsistencies in the absorption curve data provided by the solar cell manufacturers. These discrepancies make it clear that relying purely on physical equations to estimate indoor solar energy is not feasible in complex environments.

Given the limitations of theoretical models in predicting solar cell performance under real-world conditions, we turn to data-driven approaches.

\subsubsection{Performance Evaluation}

\begin{table}

\centering
\caption{ Comparison of different machine learning methods vs. LightLLM for indoor solar estimation.}
\vspace{-1em}
\label{tab:conversionresult}
\begin{tabular}{c|cc|cc}
    \toprule
    \multirow{2}{*}{\textbf{Method}} & \multicolumn{2}{c}{\textbf{Seen}} & \multicolumn{2}{c}{\textbf{Unseen}} \\ \cmidrule(lr){2-3} \cmidrule(lr){4-5}
                                     &      MSE & MAPE[\%]  &      MSE & MAPE[\%]\\
    \midrule
    Linear Regression  & 19.40 & 6.15 & 3592.97 & 168.26  \\
    Random Forest      & 5.76 & \textbf{2.58} & 2438.28 & 140.91 \\
    Gradient Boosting  & 5.50 & 2.62 & 2465.54 & 141.64 \\
    Decision Tree      & 10.29 & 2.70 & 2934.94 & 154.20 \\
    SVR               & 15.96 & 5.41 & 2013.49 & 128.54 \\
    MLP               & 6.11 & 2.68 & 819.63 & 83.02 \\
    CNN               & 5.98 & 2.60 & 959.81 & 89.61 \\
    \textbf{LightLLM}          & \textbf{5.72} & 2.59 & \textbf{263.21} & \textbf{26.33}\\
    \bottomrule
\end{tabular}
\vspace{-1em}
\end{table}

\textbf{Results Under Seen Environment. }
We then tested several machine learning methods under a seen environment, where the training set contained data from the same environment as the test set. As summarized in Table~\ref{tab:conversionresult}. In this controlled setting, data-driven methods significantly outperformed the physical equation-based approach. Among the data-driven methods, Random Forest achieved the lowest MAPE of 2.58\%, while LightLLM closely followed with a MAPE of 2.59\%. In terms of MSE, LightLLM achieved the best performance 5.72. This demonstrates that, given sufficient training data from the same environment, the data-driven models can effectively capture the relationship between spectral data and solar cell energy output.

\textbf{Results Under Unseen Environment. }
To further test the generalization of the models, we evaluated them on unseen environments, where the test set was drawn from environments not present in the training set. We specifically altered the lighting and spatial configuration within the room. For example, we introduced two diffused standing lamps, and rearranged the room layout by changing the furniture. The results for this setting are shown in Table~\ref{tab:conversionresult}. Most machine learning models suffered a significant performance drop. For example, Linear Regression and Random Forest showed MAPE values exceeding 140\%, illustrating their tendency to overfit and their lack of generalization capability. In contrast, LightLLM demonstrated remarkable robustness, achieving a MAPE of 26.33\% and MSE of 263.21, the best among the compared methods. We note that  LightLLM consistently outperforms the traditional equation-based approach in both seen and unseen environments.

\subsection{Ablation Study}
\subsubsection{LightLLM vs. Direct Prompting}

\begin{table*}[ht]
    \centering
    \caption{Performance comparison of direct prompting vs. LightLLM across models and tasks.}
    \vspace{-1em}
    \label{tab:model_comparison}
    \begin{tabular}{llcccc}
        \toprule
        \textbf{Method} & \textbf{Model} & & \textbf{Outdoor Solar Forecasting} & \textbf{Indoor Localization} & \textbf{Indoor Solar Estimation} \\
        & & & CRPS[kW]$\downarrow$ & Median Error[m]$\downarrow$ & MAPE[\%]$\downarrow$ \\
        \midrule
        \multirow{4}{*}{Zero-shot Prompting} 
        & $GPT$-4 &  & 2.57 & 1.77 & 48.72 \\
        & $GPT$-3.5 &  & 2.96 & 2.03 & 45.35 \\
        & $Llama$ 3 &  & 3.37 & 3.84 & 45.44 \\
        & $Mistral$ $Large$ 2 &  & 2.69 & 2.36 & 56.38 \\
        \midrule
        \multirow{4}{*}{Few-shot Prompting} 
        & $GPT$-4 &  & 2.42 & 1.70 & 50.43 \\
        & $GPT$-3.5 &  & 2.87 & 1.92 & 45.04 \\
        & $Llama$ 3 &  & 3.20 & 3.12 & 44.80 \\
        & $Mistral$ $Large$ 2 &  & 2.55 & 2.30 & 49.06 \\
        \midrule
        \textbf{LightLLM}
        & - &  & \textbf{1.64} & \textbf{0.98} & \textbf{26.33} \\
        \bottomrule
    \end{tabular}
    \vspace{-1em}
\end{table*}

To evaluate the effectiveness of LightLLM against a simpler method of directly prompting an LLM (e.g., ChatGPT-4), we firstly conducted an experiment using the indoor localization task as a representative. We compared both zero-shot and few-shot prompting approaches using models like GPT-4, GPT-3.5~\cite{openai2023gpt4}, Llama 3~\cite{touvron2023llama}, and Mistral Large 2~\cite{mistral2024large}. Our goal was to determine whether carefully crafted prompts, including chain-of-thought (CoT)~\cite{wei2022chain} prompting strategy, could compete with the task-specific design of LightLLM.

\textbf{Prompting Setup.} The ability of LLMs to understand and reason across a wide range of tasks makes them powerful, particularly when paired with well-constructed prompts. Recent studies~\cite{liu2023pre} have demonstrated that proper prompt engineering can significantly improve LLM performance on domain-specific tasks. Therefore, for the direct prompting method, we created a detailed prompt, leveraging CoT reasoning to improve its ability to process the sensor data step by step. A prompt example design is shown as in Figure~\ref{fig:prompt}. First, the \textbf{Instruction} gives a detailed description of the room setup, including the locations and orientations of spectral sensors, which is essential for the model to understand the spatial relationships in the environment. The \textbf{Data} section introduces the sensor readings and the changes in wavelength due to movement. The \textbf{Question} then prompts the LLM to determine the person’s location based on the given data, while the inclusion of a step-by-step analysis encourages logical reasoning, following the CoT prompting strategy.

\begin{figure}[ht]
    \centering
    \begin{tcolorbox}[colback=red!5!white, colframe=red!5!white, width=\linewidth, rounded corners,arc=5mm, boxsep=0.5mm]
        \small
        \textbf{Instruction:} In an indoor room, illuminated by overhead lights, we have placed spectral sensors all around, facing inside the room.\\
        The location of spectral sensors are as follows: \{$sensor$ $locations$ $(x, y, z)$\}.
        Their orientation are as following: \{$sensor$ $orientations$ $(x, y, z)$\}.\\
        When someone passes by, the light changes, hence the readings on each wavelength of the spectral sensors also change. When the room is empty, the sensors' readings are as follows: \{$readings$\}.\\
        \textbf{Data:} The sensor data given in the room has 18 wavelengths. The change of 18 wavelengths' reading comparing with the empty room for each sensor are given below: \{$changes$\}.\\
        \textbf{Question:} The person’s location belongs to one of the following categories: \{$locations$ $(x, y)$\}. Please tell me where the person is based on the given information. Make an analysis step by step.\\
        \textbf{Response:} \{$answer$\}
    \end{tcolorbox}
    \vspace{-1em}
    \caption{Prompt example for LSI-based localization.}
    \label{fig:prompt}
    \vspace{-1em}
\end{figure}

\textbf{Zero-shot prompting.} As shown in Table~\ref{tab:model_comparison}, GPT-4 achieved the best performance among different models, a median error of 1.77m for the LSI localization task. This error is considerably larger than the result achieved by LightLLM in the same scenario, where the median error was 0.98m. From the responses given by ChatGPT-4, we also found that, despite the detailed instruction and CoT prompt, ChatGPT-4 primarily used a very basic heuristic approach: it attempted to estimate the person's location by selecting the coordinates closest to the sensors that detected the largest change in spectral readings. 

To inject domain-specific context without additional training, researchers have explored \textbf{Few-shot prompting} by embedding a few examples directly within the prompt (e.g., ~\cite{agrawal2022large}) while the LLM parameters without tuning. From the results in Table~\ref{tab:model_comparison} we can obsereve, while this did improve the performance slightly (e.g., median error from 1.77m to 1.70m), the results still fell short compared to LightLLM.

Furthermore, we observed that certain models, such as Gemini 1.5~\cite{reid2024gemini}, refused to provide specific answers to queries about location coordinates. Instead, it offered approximate descriptions, such as ``The estimated location based on the given sensor data is near the bottom-center of the room,'' or stated, ``I cannot provide a specific coordinate.''

The limitations of direct prompting were not confined to localization. We have also conducted the experiment for the other two tasks. We used a smaller scale of the dataset to evaluate the performance of solar energy forecasting. The forecasting results by direct prompting were poor, leading to high CRPS values (e.g., GPT-4 zero-shot at 2.57 kW, compared to LightLLM's 1.64 kW). Similarly, for indoor solar energy estimation, the prompt-based estimates deviated significantly from actual measurements, as the LLMs failed to model the complexities of spectral data conversion (e.g., Llama 3 at 45.44\% vs LightLLM's 26.33\%). Interestingly, when we used few-shot prompting by providing some examples, the performance of certain models decreased slightly in indoor solar estimation task. This decline can be attributed to the fact that these examples may have misled the LLM models, causing them to overfit to specific patterns.

This ablation study shows that, LightLLM was able to handle the complexity of the sensor data much more effectively, demonstrating that for tasks requiring complex multimodal data interpretation, it is not sufficient to simply use LLM with direct prompting methods.

\subsubsection{Impact of Prompt Knowledge and LFL}
The inclusion of prompt knowledge and the LFL plays a important role in enhancing LightLLM's performance, particularly in unseen scenarios where the system must generalize to new environments not encountered during training. To demonstrate the importance of these components, we compare the performance of LightLLM with and without the prompt knowledge and LFL.

Among the tasks evaluated, localization shows the most notable improvement with prompt knowledge and LFL, in the unseen scenario, LightLLM with these components achieves a 90th percentile error of 3.46 meters, a considerable improvement over the 5.41 meters when they are removed. Table~\ref{tab:LightLLMablation} also present the results for other two tasks. For indoor solar estimation, a similar trend can be observed. The system with prompt knowledge, which includes detailed descriptions of solar cell placement and light conditions, reduces the prediction error, achieving a MAPE of 26.33\%, compared to 35.42\% without those. Similarly, we could also observe an improvement in the solar forecasting task, CRPS decrease from 2.80kW to 2.52kW.

\subsubsection{Impact of Task-specific Encoder and Low-Rank Adaptation}

\begin{table*}[ht]
\centering
\caption{Impact of LFL, TSE, and LoRA.}
\vspace{-1em}
\label{tab:LightLLMablation}
\begin{tabular}{cccccc}
    \toprule
    \multirow{2}{*}{\textbf{Task}} & \multirow{2}{*}{\textbf{Metric}} &\multicolumn{4}{c}{\textbf{Configuration}} \\ 
    & & \textbf{Normal} & \textbf{w/o LFL} & \textbf{w/o TSE} & \textbf{w/o LoRA}\\
    \midrule
    Outdoor Solar Forecasting & CRPS[kW]$\downarrow$ & \textbf{2.52} & 2.80 & 2.83 & 2.57 \\
    \midrule
    Indoor Localization & 90th Percentile Error[m]$\downarrow$ & \textbf{3.46} & 5.41 & 3.73 & 3.79 \\
    \midrule
     Indoor Solar Estimation & MAPE[\%]$\downarrow$ & \textbf{26.33} & 35.42 & 30.10 & 28.45 \\
    \bottomrule
\end{tabular}
\end{table*}

Both task-specific encoders (TSE) and LoRA are essential in affecting the performance of LightLLM across various tasks. To assess their impact, we evaluated the model’s performance with and without these components.

The results are shown in Table~\ref{tab:LightLLMablation}. For example, for the task of solar energy estimation, in unseen scenario, LightLLM with LoRA reduces the average MAPE from 28.45\% to 26.33\%. Moreover, for the solar energy forecasting task, the inclusion of a task-specific encoder brings a 12\% improvement in the CRPS metric, demonstrating its ability to capture the temporal dependencies in time-series data.

In summary, both LoRA and task-specific encoders provide considerable performance enhancement, with LoRA ensuring efficient fine-tuning for task-specific adjustments and task-specific encoders capturing the unique characteristics of each task. Together, these components make LightLLM highly accurate, flexible, and scalable across diverse applications and environments.

\subsubsection{Impact of Knowledge Graph}
In this section, we evaluate the contribution of the KG mentioned in Section~\ref{sec:KG}. In the ablation setting, we remove the KG creation in LightLLM, and instead, using raw data as input. We evaluated the models based on our LSI Localization task.

In our evaluation results, the model with KG achieved significantly lower error, with a median error of 0.98 meters. In contrast, removing the KG led to a 50\% decrease in localization accuracy, with the median error increasing to 1.42 meters. The findings validate the design rationale presented in Section~\ref{sec:KG}, demonstrating that the KG’s structured data is instrumental in achieving precise localization within the LightLLM framework.

\subsubsection{Impact of LLM in LightLLM}

\begin{figure}[t]
    \centering
    \includegraphics[width=0.85\linewidth]{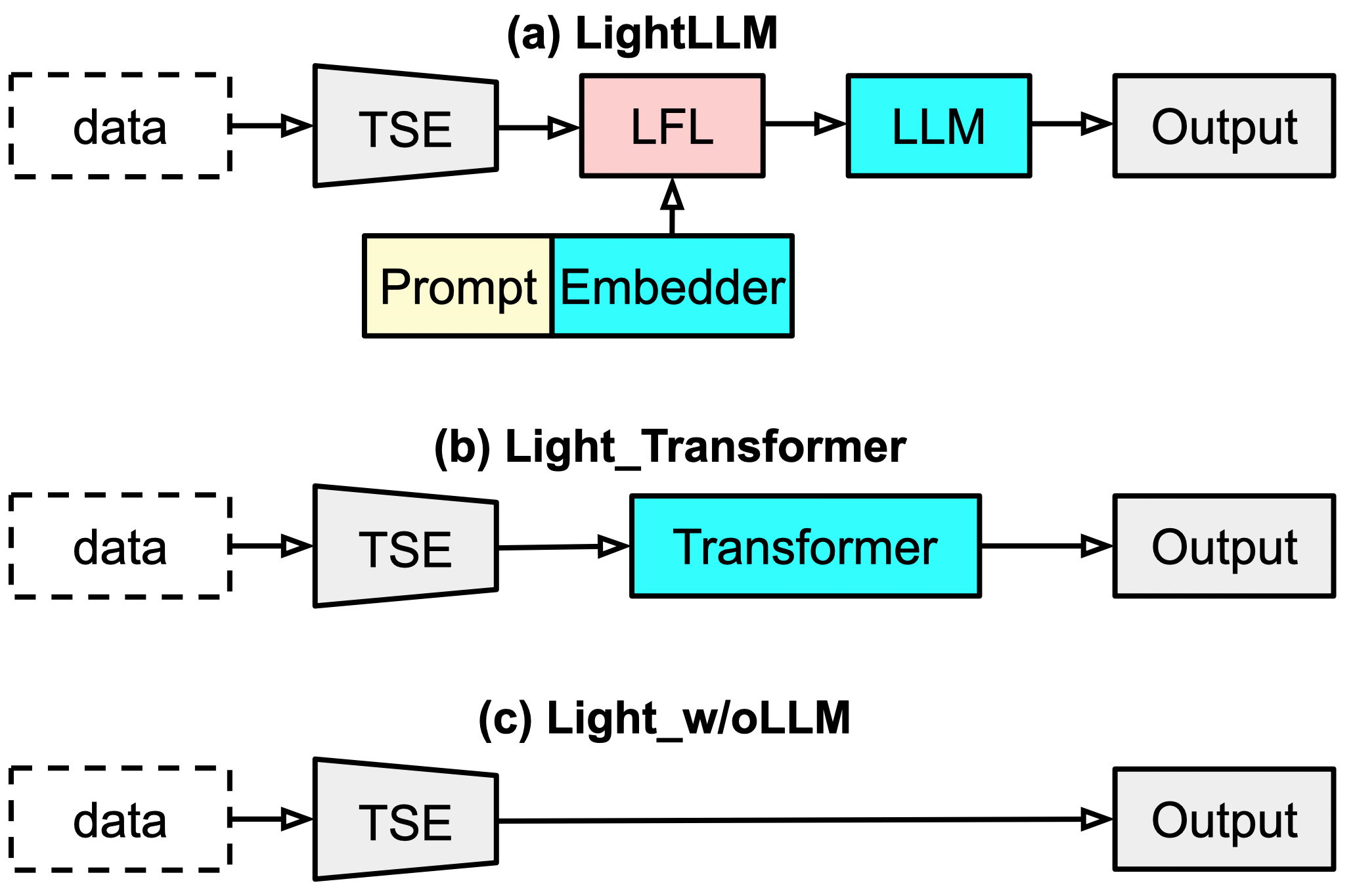}
    \caption{Three ablation versions of LightLLM.}
    \label{fig:LLMabla}
\end{figure}

\begin{figure}
    \centering
    \includegraphics[width=\linewidth]{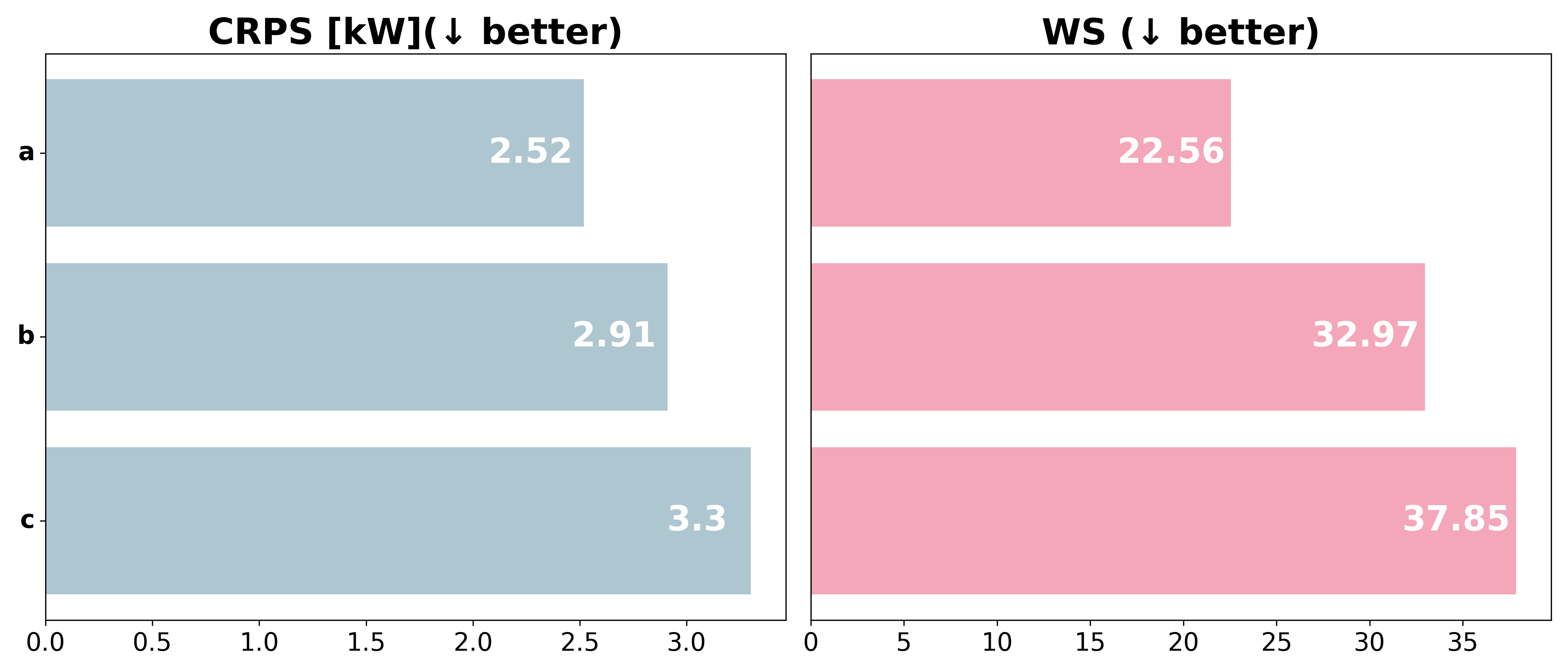}
    \caption{Performance comparison of three ablation versions of LightLLM as shown in Figure~\ref{fig:LLMabla}.}
    \label{fig:LLMuseful}
\end{figure}

To assess the importance of the LLM component within LightLLM, we conducted an ablation study by comparing three versions of our model as shown in Figure~\ref{fig:LLMabla}: LightLLM (full version), Light\_Transformer (replacing the LLM component with a single transformer), and Light\_w/oLLM (completely removing the LLM, using a simplified model). The purpose of this experiment is to evaluate whether the LLM plays a critical role in the performance of LightLLM, particularly for tasks like solar energy forecasting. The results in Figure~\ref{fig:LLMuseful} demonstrated that LightLLM (with the LLM component) consistently outperformed the other two versions across all evaluated metrics. This ablation study thus reinforces the importance of incorporating LLMs in LightLLM.

\section{FUTURE DIRECTIONS}
\label{sec:discussion}
\begin{figure}
    \centering
    \includegraphics[width=0.95\linewidth]{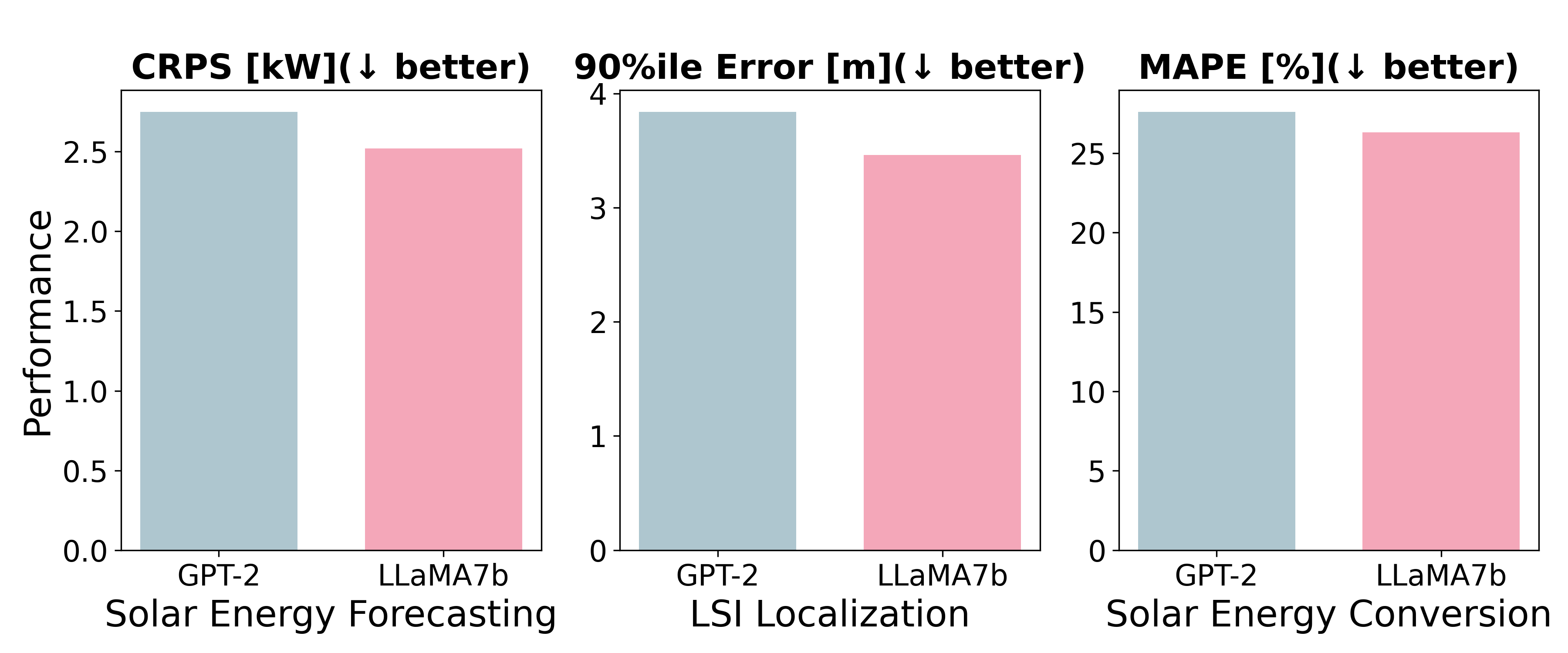}
    \vspace{-1em}
    \caption{Performance comparison between using different LLM models in LightLLM framework.}
    \label{fig:diffLLM}
    \vspace{-1em}
\end{figure}

\textbf{Exploration of Different LLM Models:} Our next attempt is to understand whether and how different pre-trained LLM models may affect the performance of LightLLM. To this end, we compared the performance of LLaMA-7B, which has 7 billion parameters, against GPT-2 with only 1.5 billion parameters across all tasks. The results in Figure~\ref{fig:diffLLM} demonstrated that LLaMA-7B consistently outperformed GPT-2.

The improvements in performance using LLaMA-7B can be attributed to its larger model size and more recent training, which have equipped it with a more extensive knowledge base and greater capacity to model complex patterns, enabling better reasoning and adaptation for spectral and temporal data.

While our experiments demonstrated the benefits of using larger models, we need to consider the trade-offs between model complexity, performance, and resource demands. Models with more parameters, like GPT-4, may further boost performance, especially for tasks requiring deeper reasoning. Future research could explore different language model architectures to identify those best suited for various PLS tasks and their impact on adaptability in complex environments.

\textbf{Task-Specific Tuning Overhead:} Although LoRA reduces the need for full model retraining, there is still some overhead involved in task-specific tuning. Fine-tuning LLMs for each new task, especially when introducing novel sensor modalities, can require additional data collection and adjustments to components such as the encoder. Future work could explore more efficient methods to streamline these processes, for example, advancing strategies to minimize data requirements and automate adaptation for emerging sensor types could make tuning faster and more resource-efficient.
\section{RELATED WORK}
\label{sec:related}

\subsection{PLS Tasks and Algorithms}
The field of PLS tasks encompasses applications like indoor localization, solar energy forecasting, and indoor solar energy estimation. These tasks are critical for optimizing energy systems, enhancing smart environments, and improving the efficiency of renewable energy grids. 

Indoor localization using light spectral information, on the other hand, is essential for services like smart navigation and asset tracking in smart buildings. Data collected from light sensors offers a promising alternative to traditional localization methods. Existing models are tailored to specific environments, and it usually relies on fingerprint collection to train the model,  which requires extensive data collection and may not generalize well to new spaces~\cite{hu2023iris,konings2019fieldlight,wang2023spectral}.

Solar radiation forecasting helps predict future energy output, improving the reliability and efficiency of energy distribution. Recent advancements in this field have largely focused on deep learning-based methods, such as combining CNN with recurrent neural network (RNN)~\cite{zhang2018deep,shi2015convolutional,paletta2021benchmarking,guen2020disentangling}. There is also a growing trend of replacing RNN with transformers to better capture temporal patterns in the data and improve forecasting accuracy~\cite{nie2024skygpt}.

With the development of efficient solar cells for indoor environments~\cite{wright2012organic,jaimes2017simple,meena2020pv}, there is a growing need for accurate indoor solar energy estimation. This task focuses on accurately predicting the energy output of photovoltaic systems in indoor conditions to maximize their utility. Current approaches often rely on the solar models that have been presented that use mathematical linear~\cite{yohanna2011model} and nonlinear functions~\cite{el2010global}.


Despite the progress made in solving these tasks, several significant challenges remain. One of the primary issues is the lack of generalization across different environments or unseen data~\cite{quinonero2022dataset}. While data augmentation~\cite{goodfellow2020generative,zhong2020random} can help improve performance in the short term, it is computationally expensive and resource-intensive, and it does not fundamentally address the issue of generalization across diverse tasks. 
Another limitation arises from the fixed nature of these models: a model designed for indoor localization cannot be easily repurposed for solar energy forecasting without re-engineering the entire architecture~\cite{weiss2016survey}. This rigidity limits the development of a scalable, multi-task solution capable of handling the diversity of data modalities in PLS tasks.

\vspace{-1em}

\subsection{Large Language Model and Applications}
In recent years,  LLMs achieved a great success in the field of natural language processing (NLP). Models like BERT~\cite{devlin2018bert}, GPT~\cite{openai2023gpt4}, and LLaMA~\cite{touvron2023llama}, which are built on the Transformer architecture, have demonstrated their ability to perform reasoning, translation and text generation tasks. One of the most significant advancements in LLMs is their ability to learn from massive datasets, capturing latent knowledge that allows them to perform well across diverse tasks.

The flexibility of LLMs has made them increasingly attractive for applications beyond NLP. Techniques like LoRA~\cite{hu2021lora} and prefix-tuning~\cite{li2021prefix} have made it possible to fine-tune LLM for specific downstream tasks with minimal computational overhead. This has led to the exploration of LLM in non-textual domains, including cross-modal learning where text is combined with images, audio, or sensor data.

Recent studies have shown the potential of LLM in multimodal tasks, where they are adapted to handle inputs from various sources beyond text. For example, models like CLIP~\cite{radford2021learning} successfully integrate image-text pairs, allowing LLM to perform visual reasoning tasks. Similarly, PaLM-E~\cite{driess2023palm} has been applied to control robotics tasks, showcasing how LLM can extend their capabilities into non-linguistic domains by integrating structured knowledge from other modalities.

In fields such as healthcare, LLM have been adapted to process clinical text alongside sensor data to provide predictions for medical outcomes. For example, medical models like PaLM-2~\cite{singhal2023towards}, MentalLLM~\cite{xu2024mental} have shown success in fine-tuning LLM to accurately interpret clinical notes and predict patient outcomes based on multimodal inputs. Additionally, NetLLM~\cite{wu2024netllm} adapts LLM for networking tasks such as viewport prediction and cluster job scheduling by leveraging the feature from networking task data, while TimeLLM~\cite{jin2023time} specializes in time-series data processing, illustrating the growing versatility of LLM across diverse domains. Similarly, FM-Fi~\cite{weng2024large} explores LLM applications in human activity recognition (HAR) using RF data. These advancements illustrate the cross-domain adaptability of LLMs, inspiring the development of LightLLM for PLS tasks.


\section{CONCLUSION}
\label{sec:conclusion}

We presented LightLLM, a unified framework designed to address diverse PLS tasks such as indoor localization, outdoor solar forecasting, and indoor solar estimation. The Latent Fusion Layer and task-specific prompts enable effective multimodal sensor fusion, providing a rich context information. Moreover, by integrating LLM with task-specific encoders and utilizing techniques like LoRA for efficient fine-tuning, LightLLM achieved significant improvements in accuracy and adaptability, particularly in unseen scenarios. Our extensive evaluations showed that LightLLM outperforms traditional methods and and simple prompting approaches, in handling complex, real-world data. Though evaluated on different tasks, LightLLM’s design offers a general-purpose solution for other context-rich sensing applications, establishing a foundation for LLM-based frameworks in PLS.

\begin{acks}
This work was partly supported by the Australian Research Council Discovery Project DP210100904 and CSIRO Data61 PhD Scholarship Program.
\end{acks}

\bibliographystyle{ACM-Reference-Format}


\begin{thebibliography}{57}


\ifx \showCODEN    \undefined \def \showCODEN     #1{\unskip}     \fi
\ifx \showDOI      \undefined \def \showDOI       #1{#1}\fi
\ifx \showISBNx    \undefined \def \showISBNx     #1{\unskip}     \fi
\ifx \showISBNxiii \undefined \def \showISBNxiii  #1{\unskip}     \fi
\ifx \showISSN     \undefined \def \showISSN      #1{\unskip}     \fi
\ifx \showLCCN     \undefined \def \showLCCN      #1{\unskip}     \fi
\ifx \shownote     \undefined \def \shownote      #1{#1}          \fi
\ifx \showarticletitle \undefined \def \showarticletitle #1{#1}   \fi
\ifx \showURL      \undefined \def \showURL       {\relax}        \fi
\providecommand\bibfield[2]{#2}
\providecommand\bibinfo[2]{#2}
\providecommand\natexlab[1]{#1}
\providecommand\showeprint[2][]{arXiv:#2}

\bibitem[sam(2023)]%
        {samsungTV}
 \bibinfo{year}{2023}\natexlab{}.
\newblock \bibinfo{title}{Original Samsung Smart TV Remote (Solar Cell Remote Control - BN59-01385B Genuine)}.
\newblock \bibinfo{howpublished}{\url{https://remotecontrolwarehouse.com.au/products/}}.
\newblock


\bibitem[Agrawal et~al\mbox{.}(2022)]%
        {agrawal2022large}
\bibfield{author}{\bibinfo{person}{Monica Agrawal}, \bibinfo{person}{Stefan Hegselmann}, \bibinfo{person}{Hunter Lang}, \bibinfo{person}{Yoon Kim}, {and} \bibinfo{person}{David Sontag}.} \bibinfo{year}{2022}\natexlab{}.
\newblock \showarticletitle{Large language models are few-shot clinical information extractors}.
\newblock \bibinfo{journal}{\emph{arXiv preprint arXiv:2205.12689}} (\bibinfo{year}{2022}).
\newblock


\bibitem[AI(2024)]%
        {mistral2024large}
\bibfield{author}{\bibinfo{person}{Mistral AI}.} \bibinfo{year}{2024}\natexlab{}.
\newblock \bibinfo{title}{Mistral}.
\newblock \bibinfo{howpublished}{\url{https://mistral.ai/}}.
\newblock


\bibitem[ams OSRAM(2022)]%
        {as7265x}
\bibfield{author}{\bibinfo{person}{ams OSRAM}.} \bibinfo{year}{2022}\natexlab{}.
\newblock \bibinfo{title}{{AS7265x Smart Spectral Sensor}}.
\newblock \bibinfo{howpublished}{\url{https://ams.com/en/as7265x}}.
\newblock


\bibitem[Balaji et~al\mbox{.}(2016)]%
        {balaji2016brick}
\bibfield{author}{\bibinfo{person}{Bharathan Balaji}, \bibinfo{person}{Arka Bhattacharya}, \bibinfo{person}{Gabriel Fierro}, \bibinfo{person}{Jingkun Gao}, \bibinfo{person}{Joshua Gluck}, \bibinfo{person}{Dezhi Hong}, \bibinfo{person}{Aslak Johansen}, \bibinfo{person}{Jason Koh}, \bibinfo{person}{Joern Ploennigs}, \bibinfo{person}{Yuvraj Agarwal}, {et~al\mbox{.}}} \bibinfo{year}{2016}\natexlab{}.
\newblock \showarticletitle{Brick: Towards a unified metadata schema for buildings}. In \bibinfo{booktitle}{\emph{Proceedings of the 3rd ACM International Conference on Systems for Energy-Efficient Built Environments}}. \bibinfo{pages}{41--50}.
\newblock


\bibitem[Benke and Tomkins(2017)]%
        {benke2017future}
\bibfield{author}{\bibinfo{person}{Kurt Benke} {and} \bibinfo{person}{Bruce Tomkins}.} \bibinfo{year}{2017}\natexlab{}.
\newblock \showarticletitle{Future food-production systems: vertical farming and controlled-environment agriculture}.
\newblock \bibinfo{journal}{\emph{Sustainability: Science, Practice and Policy}} \bibinfo{volume}{13}, \bibinfo{number}{1} (\bibinfo{year}{2017}), \bibinfo{pages}{13--26}.
\newblock


\bibitem[Biswas and Kim(2020)]%
        {biswas2020solar}
\bibfield{author}{\bibinfo{person}{Swarup Biswas} {and} \bibinfo{person}{Hyeok Kim}.} \bibinfo{year}{2020}\natexlab{}.
\newblock \showarticletitle{Solar cells for indoor applications: Progress and development}.
\newblock \bibinfo{journal}{\emph{Polymers}} \bibinfo{volume}{12}, \bibinfo{number}{6} (\bibinfo{year}{2020}), \bibinfo{pages}{1338}.
\newblock


\bibitem[Blagec et~al\mbox{.}(2022)]%
        {blagec2022curated}
\bibfield{author}{\bibinfo{person}{Kathrin Blagec}, \bibinfo{person}{Adriano Barbosa-Silva}, \bibinfo{person}{Simon Ott}, {and} \bibinfo{person}{Matthias Samwald}.} \bibinfo{year}{2022}\natexlab{}.
\newblock \showarticletitle{A curated, ontology-based, large-scale knowledge graph of artificial intelligence tasks and benchmarks}.
\newblock \bibinfo{journal}{\emph{Scientific Data}} \bibinfo{volume}{9}, \bibinfo{number}{1} (\bibinfo{year}{2022}), \bibinfo{pages}{322}.
\newblock


\bibitem[Da~Rosa and Ordonez(2021)]%
        {da2021fundamentals}
\bibfield{author}{\bibinfo{person}{Aldo~Vieira Da~Rosa} {and} \bibinfo{person}{Juan~Carlos Ordonez}.} \bibinfo{year}{2021}\natexlab{}.
\newblock \bibinfo{booktitle}{\emph{Fundamentals of renewable energy processes}}.
\newblock \bibinfo{publisher}{Academic Press}.
\newblock


\bibitem[Devlin(2018)]%
        {devlin2018bert}
\bibfield{author}{\bibinfo{person}{Jacob Devlin}.} \bibinfo{year}{2018}\natexlab{}.
\newblock \showarticletitle{Bert: Pre-training of deep bidirectional transformers for language understanding}.
\newblock \bibinfo{journal}{\emph{arXiv preprint arXiv:1810.04805}} (\bibinfo{year}{2018}).
\newblock


\bibitem[Driess et~al\mbox{.}(2023)]%
        {driess2023palm}
\bibfield{author}{\bibinfo{person}{Danny Driess}, \bibinfo{person}{Fei Xia}, \bibinfo{person}{Mehdi~SM Sajjadi}, \bibinfo{person}{Corey Lynch}, \bibinfo{person}{Aakanksha Chowdhery}, \bibinfo{person}{Brian Ichter}, \bibinfo{person}{Ayzaan Wahid}, \bibinfo{person}{Jonathan Tompson}, \bibinfo{person}{Quan Vuong}, \bibinfo{person}{Tianhe Yu}, {et~al\mbox{.}}} \bibinfo{year}{2023}\natexlab{}.
\newblock \showarticletitle{Palm-e: An embodied multimodal language model}.
\newblock \bibinfo{journal}{\emph{arXiv preprint arXiv:2303.03378}} (\bibinfo{year}{2023}).
\newblock


\bibitem[El-Sebaii et~al\mbox{.}(2010)]%
        {el2010global}
\bibfield{author}{\bibinfo{person}{AA El-Sebaii}, \bibinfo{person}{FS Al-Hazmi}, \bibinfo{person}{AA Al-Ghamdi}, {and} \bibinfo{person}{Saud~Jameel Yaghmour}.} \bibinfo{year}{2010}\natexlab{}.
\newblock \showarticletitle{Global, direct and diffuse solar radiation on horizontal and tilted surfaces in Jeddah, Saudi Arabia}.
\newblock \bibinfo{journal}{\emph{Applied energy}} \bibinfo{volume}{87}, \bibinfo{number}{2} (\bibinfo{year}{2010}), \bibinfo{pages}{568--576}.
\newblock


\bibitem[Gneiting and Katzfuss(2014)]%
        {gneiting2014probabilistic}
\bibfield{author}{\bibinfo{person}{Tilmann Gneiting} {and} \bibinfo{person}{Matthias Katzfuss}.} \bibinfo{year}{2014}\natexlab{}.
\newblock \showarticletitle{Probabilistic forecasting}.
\newblock \bibinfo{journal}{\emph{Annual Review of Statistics and Its Application}} \bibinfo{volume}{1}, \bibinfo{number}{1} (\bibinfo{year}{2014}), \bibinfo{pages}{125--151}.
\newblock


\bibitem[Goodfellow(2016)]%
        {goodfellow2016deep}
\bibfield{author}{\bibinfo{person}{Ian Goodfellow}.} \bibinfo{year}{2016}\natexlab{}.
\newblock \bibinfo{title}{Deep learning}.
\newblock
\newblock


\bibitem[Goodfellow et~al\mbox{.}(2020)]%
        {goodfellow2020generative}
\bibfield{author}{\bibinfo{person}{Ian Goodfellow}, \bibinfo{person}{Jean Pouget-Abadie}, \bibinfo{person}{Mehdi Mirza}, \bibinfo{person}{Bing Xu}, \bibinfo{person}{David Warde-Farley}, \bibinfo{person}{Sherjil Ozair}, \bibinfo{person}{Aaron Courville}, {and} \bibinfo{person}{Yoshua Bengio}.} \bibinfo{year}{2020}\natexlab{}.
\newblock \showarticletitle{Generative adversarial networks}.
\newblock \bibinfo{journal}{\emph{Commun. ACM}} \bibinfo{volume}{63}, \bibinfo{number}{11} (\bibinfo{year}{2020}), \bibinfo{pages}{139--144}.
\newblock


\bibitem[Guen and Thome(2020)]%
        {guen2020disentangling}
\bibfield{author}{\bibinfo{person}{Vincent~Le Guen} {and} \bibinfo{person}{Nicolas Thome}.} \bibinfo{year}{2020}\natexlab{}.
\newblock \showarticletitle{Disentangling physical dynamics from unknown factors for unsupervised video prediction}. In \bibinfo{booktitle}{\emph{Proceedings of the IEEE/CVF conference on computer vision and pattern recognition}}. \bibinfo{pages}{11474--11484}.
\newblock


\bibitem[Hu et~al\mbox{.}(2021)]%
        {hu2021lora}
\bibfield{author}{\bibinfo{person}{Edward~J Hu}, \bibinfo{person}{Yelong Shen}, \bibinfo{person}{Phillip Wallis}, \bibinfo{person}{Zeyuan Allen-Zhu}, \bibinfo{person}{Yuanzhi Li}, \bibinfo{person}{Shean Wang}, \bibinfo{person}{Lu Wang}, {and} \bibinfo{person}{Weizhu Chen}.} \bibinfo{year}{2021}\natexlab{}.
\newblock \showarticletitle{Lora: Low-rank adaptation of large language models}.
\newblock \bibinfo{journal}{\emph{arXiv preprint arXiv:2106.09685}} (\bibinfo{year}{2021}).
\newblock


\bibitem[Hu et~al\mbox{.}(2023)]%
        {hu2023iris}
\bibfield{author}{\bibinfo{person}{Jiawei Hu}, \bibinfo{person}{Yanxiang Wang}, \bibinfo{person}{Hong Jia}, \bibinfo{person}{Wen Hu}, \bibinfo{person}{Mahbub Hassan}, \bibinfo{person}{Brano Kusy}, \bibinfo{person}{Ashraf Uddin}, {and} \bibinfo{person}{Moustafa Youssef}.} \bibinfo{year}{2023}\natexlab{}.
\newblock \showarticletitle{Iris: Passive Visible Light Positioning Using Light Spectral Information}.
\newblock \bibinfo{journal}{\emph{Proceedings of the ACM on Interactive, Mobile, Wearable and Ubiquitous Technologies}} \bibinfo{volume}{7}, \bibinfo{number}{3} (\bibinfo{year}{2023}), \bibinfo{pages}{1--27}.
\newblock


\bibitem[Hu et~al\mbox{.}(2024)]%
        {hu2024lidarspectra}
\bibfield{author}{\bibinfo{person}{Jiawei Hu}, \bibinfo{person}{Yanxiang Wang}, \bibinfo{person}{Hong Jia}, \bibinfo{person}{Cheng Jiang}, \bibinfo{person}{Mahbub Hassan}, \bibinfo{person}{Brano Kusy}, {and} \bibinfo{person}{Wen Hu}.} \bibinfo{year}{2024}\natexlab{}.
\newblock \showarticletitle{LiDARSpectra: Synthetic Indoor Spectral Mapping with Low-cost LiDARs}. In \bibinfo{booktitle}{\emph{2024 23rd ACM/IEEE International Conference on Information Processing in Sensor Networks (IPSN)}}. IEEE, \bibinfo{pages}{75--87}.
\newblock


\bibitem[Inman et~al\mbox{.}(2013)]%
        {inman2013solar}
\bibfield{author}{\bibinfo{person}{Rich~H Inman}, \bibinfo{person}{Hugo~TC Pedro}, {and} \bibinfo{person}{Carlos~FM Coimbra}.} \bibinfo{year}{2013}\natexlab{}.
\newblock \showarticletitle{Solar forecasting methods for renewable energy integration}.
\newblock \bibinfo{journal}{\emph{Progress in energy and combustion science}} \bibinfo{volume}{39}, \bibinfo{number}{6} (\bibinfo{year}{2013}), \bibinfo{pages}{535--576}.
\newblock


\bibitem[Jaimes and de~Sousa(2017)]%
        {jaimes2017simple}
\bibfield{author}{\bibinfo{person}{Arturo~Fajardo Jaimes} {and} \bibinfo{person}{Fernando~Rangel de Sousa}.} \bibinfo{year}{2017}\natexlab{}.
\newblock \showarticletitle{Simple modeling of photovoltaic solar cells for indoor harvesting applications}.
\newblock \bibinfo{journal}{\emph{Solar Energy}}  \bibinfo{volume}{157} (\bibinfo{year}{2017}), \bibinfo{pages}{792--802}.
\newblock


\bibitem[Ji et~al\mbox{.}(2024)]%
        {ji2024hargpt}
\bibfield{author}{\bibinfo{person}{Sijie Ji}, \bibinfo{person}{Xinzhe Zheng}, {and} \bibinfo{person}{Chenshu Wu}.} \bibinfo{year}{2024}\natexlab{}.
\newblock \showarticletitle{HARGPT: Are LLMs Zero-Shot Human Activity Recognizers?}
\newblock \bibinfo{journal}{\emph{arXiv preprint arXiv:2403.02727}} (\bibinfo{year}{2024}).
\newblock


\bibitem[Jin et~al\mbox{.}(2023)]%
        {jin2023time}
\bibfield{author}{\bibinfo{person}{Ming Jin}, \bibinfo{person}{Shiyu Wang}, \bibinfo{person}{Lintao Ma}, \bibinfo{person}{Zhixuan Chu}, \bibinfo{person}{James~Y Zhang}, \bibinfo{person}{Xiaoming Shi}, \bibinfo{person}{Pin-Yu Chen}, \bibinfo{person}{Yuxuan Liang}, \bibinfo{person}{Yuan-Fang Li}, \bibinfo{person}{Shirui Pan}, {et~al\mbox{.}}} \bibinfo{year}{2023}\natexlab{}.
\newblock \showarticletitle{Time-llm: Time series forecasting by reprogramming large language models}.
\newblock \bibinfo{journal}{\emph{arXiv preprint arXiv:2310.01728}} (\bibinfo{year}{2023}).
\newblock


\bibitem[Konings et~al\mbox{.}(2019)]%
        {konings2019fieldlight}
\bibfield{author}{\bibinfo{person}{Daniel Konings}, \bibinfo{person}{Nathaniel Faulkner}, \bibinfo{person}{Fakhrul Alam}, \bibinfo{person}{Edmund M-K Lai}, {and} \bibinfo{person}{Serge Demidenko}.} \bibinfo{year}{2019}\natexlab{}.
\newblock \showarticletitle{FieldLight: Device-free indoor human localization using passive visible light positioning and artificial potential fields}.
\newblock \bibinfo{journal}{\emph{IEEE Sensors Journal}} \bibinfo{volume}{20}, \bibinfo{number}{2} (\bibinfo{year}{2019}), \bibinfo{pages}{1054--1066}.
\newblock


\bibitem[Lea et~al\mbox{.}(2017)]%
        {lea2017temporal}
\bibfield{author}{\bibinfo{person}{Colin Lea}, \bibinfo{person}{Michael~D Flynn}, \bibinfo{person}{Rene Vidal}, \bibinfo{person}{Austin Reiter}, {and} \bibinfo{person}{Gregory~D Hager}.} \bibinfo{year}{2017}\natexlab{}.
\newblock \showarticletitle{Temporal convolutional networks for action segmentation and detection}. In \bibinfo{booktitle}{\emph{proceedings of the IEEE Conference on Computer Vision and Pattern Recognition}}. \bibinfo{pages}{156--165}.
\newblock


\bibitem[Li and Liang(2021)]%
        {li2021prefix}
\bibfield{author}{\bibinfo{person}{Xiang~Lisa Li} {and} \bibinfo{person}{Percy Liang}.} \bibinfo{year}{2021}\natexlab{}.
\newblock \showarticletitle{Prefix-tuning: Optimizing continuous prompts for generation}.
\newblock \bibinfo{journal}{\emph{arXiv preprint arXiv:2101.00190}} (\bibinfo{year}{2021}).
\newblock


\bibitem[Li et~al\mbox{.}(2021)]%
        {li2021survey}
\bibfield{author}{\bibinfo{person}{Zewen Li}, \bibinfo{person}{Fan Liu}, \bibinfo{person}{Wenjie Yang}, \bibinfo{person}{Shouheng Peng}, {and} \bibinfo{person}{Jun Zhou}.} \bibinfo{year}{2021}\natexlab{}.
\newblock \showarticletitle{A survey of convolutional neural networks: analysis, applications, and prospects}.
\newblock \bibinfo{journal}{\emph{IEEE transactions on neural networks and learning systems}} \bibinfo{volume}{33}, \bibinfo{number}{12} (\bibinfo{year}{2021}), \bibinfo{pages}{6999--7019}.
\newblock


\bibitem[Liu et~al\mbox{.}(2023)]%
        {liu2023pre}
\bibfield{author}{\bibinfo{person}{Pengfei Liu}, \bibinfo{person}{Weizhe Yuan}, \bibinfo{person}{Jinlan Fu}, \bibinfo{person}{Zhengbao Jiang}, \bibinfo{person}{Hiroaki Hayashi}, {and} \bibinfo{person}{Graham Neubig}.} \bibinfo{year}{2023}\natexlab{}.
\newblock \showarticletitle{Pre-train, prompt, and predict: A systematic survey of prompting methods in natural language processing}.
\newblock \bibinfo{journal}{\emph{Comput. Surveys}} \bibinfo{volume}{55}, \bibinfo{number}{9} (\bibinfo{year}{2023}), \bibinfo{pages}{1--35}.
\newblock


\bibitem[Ma et~al\mbox{.}(2019)]%
        {ma2019solargest}
\bibfield{author}{\bibinfo{person}{Dong Ma}, \bibinfo{person}{Guohao Lan}, \bibinfo{person}{Mahbub Hassan}, \bibinfo{person}{Wen Hu}, \bibinfo{person}{Mushfika~B Upama}, \bibinfo{person}{Ashraf Uddin}, {and} \bibinfo{person}{Moustafa Youssef}.} \bibinfo{year}{2019}\natexlab{}.
\newblock \showarticletitle{Solargest: Ubiquitous and battery-free gesture recognition using solar cells}. In \bibinfo{booktitle}{\emph{The 25th annual international conference on mobile computing and networking}}. \bibinfo{pages}{1--15}.
\newblock


\bibitem[Meena et~al\mbox{.}(2020)]%
        {meena2020pv}
\bibfield{author}{\bibinfo{person}{Yogesh~Kumar Meena}, \bibinfo{person}{Krishna Seunarine}, \bibinfo{person}{Deepak~Ranjan Sahoo}, \bibinfo{person}{Simon Robinson}, \bibinfo{person}{Jennifer Pearson}, \bibinfo{person}{Chi Zhang}, \bibinfo{person}{Matt Carnie}, \bibinfo{person}{Adam Pockett}, \bibinfo{person}{Andrew Prescott}, \bibinfo{person}{Suzanne~K Thomas}, {et~al\mbox{.}}} \bibinfo{year}{2020}\natexlab{}.
\newblock \showarticletitle{PV-tiles: towards closely-coupled photovoltaic and digital materials for useful, beautiful and sustainable interactive surfaces}. In \bibinfo{booktitle}{\emph{Proceedings of the 2020 CHI Conference on Human Factors in Computing Systems}}. \bibinfo{pages}{1--12}.
\newblock


\bibitem[Nie et~al\mbox{.}(2023)]%
        {nie2023skipp}
\bibfield{author}{\bibinfo{person}{Yuhao Nie}, \bibinfo{person}{Xiatong Li}, \bibinfo{person}{Andea Scott}, \bibinfo{person}{Yuchi Sun}, \bibinfo{person}{Vignesh Venugopal}, {and} \bibinfo{person}{Adam Brandt}.} \bibinfo{year}{2023}\natexlab{}.
\newblock \showarticletitle{SKIPP’D: A SKy Images and Photovoltaic Power Generation Dataset for short-term solar forecasting}.
\newblock \bibinfo{journal}{\emph{Solar Energy}}  \bibinfo{volume}{255} (\bibinfo{year}{2023}), \bibinfo{pages}{171--179}.
\newblock


\bibitem[Nie et~al\mbox{.}(2024)]%
        {nie2024skygpt}
\bibfield{author}{\bibinfo{person}{Yuhao Nie}, \bibinfo{person}{Eric Zelikman}, \bibinfo{person}{Andea Scott}, \bibinfo{person}{Quentin Paletta}, {and} \bibinfo{person}{Adam Brandt}.} \bibinfo{year}{2024}\natexlab{}.
\newblock \showarticletitle{SkyGPT: Probabilistic ultra-short-term solar forecasting using synthetic sky images from physics-constrained VideoGPT}.
\newblock \bibinfo{journal}{\emph{Advances in Applied Energy}}  \bibinfo{volume}{14} (\bibinfo{year}{2024}), \bibinfo{pages}{100172}.
\newblock


\bibitem[OpenAI(2023)]%
        {openai2023gpt4}
\bibfield{author}{\bibinfo{person}{OpenAI}.} \bibinfo{year}{2023}\natexlab{}.
\newblock \bibinfo{title}{GPT-4 Technical Report}.
\newblock \bibinfo{howpublished}{\url{https://openai.com/research/gpt-4}}.
\newblock


\bibitem[Paletta et~al\mbox{.}(2021)]%
        {paletta2021benchmarking}
\bibfield{author}{\bibinfo{person}{Quentin Paletta}, \bibinfo{person}{Guillaume Arbod}, {and} \bibinfo{person}{Joan Lasenby}.} \bibinfo{year}{2021}\natexlab{}.
\newblock \showarticletitle{Benchmarking of deep learning irradiance forecasting models from sky images--An in-depth analysis}.
\newblock \bibinfo{journal}{\emph{Solar Energy}}  \bibinfo{volume}{224} (\bibinfo{year}{2021}), \bibinfo{pages}{855--867}.
\newblock


\bibitem[Parretta et~al\mbox{.}(1998)]%
        {parretta1998effects}
\bibfield{author}{\bibinfo{person}{Antonio Parretta}, \bibinfo{person}{Angelo Sarno}, {and} \bibinfo{person}{Luciano~RM Vicari}.} \bibinfo{year}{1998}\natexlab{}.
\newblock \showarticletitle{Effects of solar irradiation conditions on the outdoor performance of photovoltaic modules}.
\newblock \bibinfo{journal}{\emph{Optics Communications}} \bibinfo{volume}{153}, \bibinfo{number}{1-3} (\bibinfo{year}{1998}), \bibinfo{pages}{153--163}.
\newblock


\bibitem[Qui{\~n}onero-Candela et~al\mbox{.}(2022)]%
        {quinonero2022dataset}
\bibfield{author}{\bibinfo{person}{Joaquin Qui{\~n}onero-Candela}, \bibinfo{person}{Masashi Sugiyama}, \bibinfo{person}{Anton Schwaighofer}, {and} \bibinfo{person}{Neil~D Lawrence}.} \bibinfo{year}{2022}\natexlab{}.
\newblock \bibinfo{booktitle}{\emph{Dataset shift in machine learning}}.
\newblock \bibinfo{publisher}{Mit Press}.
\newblock


\bibitem[Radford et~al\mbox{.}(2021)]%
        {radford2021learning}
\bibfield{author}{\bibinfo{person}{Alec Radford}, \bibinfo{person}{Jong~Wook Kim}, \bibinfo{person}{Chris Hallacy}, \bibinfo{person}{Aditya Ramesh}, \bibinfo{person}{Gabriel Goh}, \bibinfo{person}{Sandhini Agarwal}, \bibinfo{person}{Girish Sastry}, \bibinfo{person}{Amanda Askell}, \bibinfo{person}{Pamela Mishkin}, \bibinfo{person}{Jack Clark}, {et~al\mbox{.}}} \bibinfo{year}{2021}\natexlab{}.
\newblock \showarticletitle{Learning transferable visual models from natural language supervision}. In \bibinfo{booktitle}{\emph{International conference on machine learning}}. PMLR, \bibinfo{pages}{8748--8763}.
\newblock


\bibitem[Reid et~al\mbox{.}(2024)]%
        {reid2024gemini}
\bibfield{author}{\bibinfo{person}{Machel Reid}, \bibinfo{person}{Nikolay Savinov}, \bibinfo{person}{Denis Teplyashin}, \bibinfo{person}{Dmitry Lepikhin}, \bibinfo{person}{Timothy Lillicrap}, \bibinfo{person}{Jean-baptiste Alayrac}, \bibinfo{person}{Radu Soricut}, \bibinfo{person}{Angeliki Lazaridou}, \bibinfo{person}{Orhan Firat}, \bibinfo{person}{Julian Schrittwieser}, {et~al\mbox{.}}} \bibinfo{year}{2024}\natexlab{}.
\newblock \showarticletitle{Gemini 1.5: Unlocking multimodal understanding across millions of tokens of context}.
\newblock \bibinfo{journal}{\emph{arXiv preprint arXiv:2403.05530}} (\bibinfo{year}{2024}).
\newblock


\bibitem[Shi et~al\mbox{.}(2015)]%
        {shi2015convolutional}
\bibfield{author}{\bibinfo{person}{Xingjian Shi}, \bibinfo{person}{Zhourong Chen}, \bibinfo{person}{Hao Wang}, \bibinfo{person}{Dit-Yan Yeung}, \bibinfo{person}{Wai-Kin Wong}, {and} \bibinfo{person}{Wang-chun Woo}.} \bibinfo{year}{2015}\natexlab{}.
\newblock \showarticletitle{Convolutional LSTM network: A machine learning approach for precipitation nowcasting}.
\newblock \bibinfo{journal}{\emph{Advances in neural information processing systems}}  \bibinfo{volume}{28} (\bibinfo{year}{2015}).
\newblock


\bibitem[Singhal et~al\mbox{.}(2023)]%
        {singhal2023towards}
\bibfield{author}{\bibinfo{person}{Karan Singhal}, \bibinfo{person}{Tao Tu}, \bibinfo{person}{Juraj Gottweis}, \bibinfo{person}{Rory Sayres}, \bibinfo{person}{Ellery Wulczyn}, \bibinfo{person}{Le Hou}, \bibinfo{person}{Kevin Clark}, \bibinfo{person}{Stephen Pfohl}, \bibinfo{person}{Heather Cole-Lewis}, \bibinfo{person}{Darlene Neal}, {et~al\mbox{.}}} \bibinfo{year}{2023}\natexlab{}.
\newblock \showarticletitle{Towards expert-level medical question answering with large language models}.
\newblock \bibinfo{journal}{\emph{arXiv preprint arXiv:2305.09617}} (\bibinfo{year}{2023}).
\newblock


\bibitem[Touvron et~al\mbox{.}(2023)]%
        {touvron2023llama}
\bibfield{author}{\bibinfo{person}{Hugo Touvron}, \bibinfo{person}{Thibaut Lavril}, \bibinfo{person}{Gautier Izacard}, \bibinfo{person}{Xavier Martinet}, \bibinfo{person}{Marie-Anne Lachaux}, \bibinfo{person}{Timoth{\'e}e Lacroix}, \bibinfo{person}{Baptiste Rozi{\`e}re}, \bibinfo{person}{Naman Goyal}, \bibinfo{person}{Eric Hambro}, \bibinfo{person}{Faisal Azhar}, {et~al\mbox{.}}} \bibinfo{year}{2023}\natexlab{}.
\newblock \showarticletitle{Llama: Open and efficient foundation language models}.
\newblock \bibinfo{journal}{\emph{arXiv preprint arXiv:2302.13971}} (\bibinfo{year}{2023}).
\newblock


\bibitem[Vaswani(2017)]%
        {vaswani2017attention}
\bibfield{author}{\bibinfo{person}{A Vaswani}.} \bibinfo{year}{2017}\natexlab{}.
\newblock \showarticletitle{Attention is all you need}.
\newblock \bibinfo{journal}{\emph{Advances in Neural Information Processing Systems}} (\bibinfo{year}{2017}).
\newblock


\bibitem[Wang et~al\mbox{.}(2023)]%
        {wang2023spectral}
\bibfield{author}{\bibinfo{person}{Yanxiang Wang}, \bibinfo{person}{Jiawei Hu}, \bibinfo{person}{Hong Jia}, \bibinfo{person}{Wen Hu}, \bibinfo{person}{Mahbub Hassan}, \bibinfo{person}{Ashraf Uddin}, \bibinfo{person}{Brano Kusy}, {and} \bibinfo{person}{Moustafa Youssef}.} \bibinfo{year}{2023}\natexlab{}.
\newblock \showarticletitle{Spectral-Loc: Indoor Localization using Light Spectral Information}.
\newblock \bibinfo{journal}{\emph{Proceedings of the ACM on Interactive, Mobile, Wearable and Ubiquitous Technologies}} \bibinfo{volume}{7}, \bibinfo{number}{1} (\bibinfo{year}{2023}), \bibinfo{pages}{1--26}.
\newblock


\bibitem[Wang and Srinivasan(2017)]%
        {wang2017review}
\bibfield{author}{\bibinfo{person}{Zeyu Wang} {and} \bibinfo{person}{Ravi~S Srinivasan}.} \bibinfo{year}{2017}\natexlab{}.
\newblock \showarticletitle{A review of artificial intelligence based building energy use prediction: Contrasting the capabilities of single and ensemble prediction models}.
\newblock \bibinfo{journal}{\emph{Renewable and Sustainable Energy Reviews}}  \bibinfo{volume}{75} (\bibinfo{year}{2017}), \bibinfo{pages}{796--808}.
\newblock


\bibitem[Wei et~al\mbox{.}(2022)]%
        {wei2022chain}
\bibfield{author}{\bibinfo{person}{Jason Wei}, \bibinfo{person}{Xuezhi Wang}, \bibinfo{person}{Dale Schuurmans}, \bibinfo{person}{Maarten Bosma}, \bibinfo{person}{Fei Xia}, \bibinfo{person}{Ed Chi}, \bibinfo{person}{Quoc~V Le}, \bibinfo{person}{Denny Zhou}, {et~al\mbox{.}}} \bibinfo{year}{2022}\natexlab{}.
\newblock \showarticletitle{Chain-of-thought prompting elicits reasoning in large language models}.
\newblock \bibinfo{journal}{\emph{Advances in neural information processing systems}}  \bibinfo{volume}{35} (\bibinfo{year}{2022}), \bibinfo{pages}{24824--24837}.
\newblock


\bibitem[Weiss et~al\mbox{.}(2016)]%
        {weiss2016survey}
\bibfield{author}{\bibinfo{person}{Karl Weiss}, \bibinfo{person}{Taghi~M Khoshgoftaar}, {and} \bibinfo{person}{DingDing Wang}.} \bibinfo{year}{2016}\natexlab{}.
\newblock \showarticletitle{A survey of transfer learning}.
\newblock \bibinfo{journal}{\emph{Journal of Big data}}  \bibinfo{volume}{3} (\bibinfo{year}{2016}), \bibinfo{pages}{1--40}.
\newblock


\bibitem[Weng et~al\mbox{.}(2024)]%
        {weng2024large}
\bibfield{author}{\bibinfo{person}{Yuxuan Weng}, \bibinfo{person}{Guoquan Wu}, \bibinfo{person}{Tianyue Zheng}, \bibinfo{person}{Yanbing Yang}, {and} \bibinfo{person}{Jun Luo}.} \bibinfo{year}{2024}\natexlab{}.
\newblock \showarticletitle{Large Model for Small Data: Foundation Model for Cross-Modal RF Human Activity Recognition}. In \bibinfo{booktitle}{\emph{Proceedings of the 22nd ACM Conference on Embedded Networked Sensor Systems}}. \bibinfo{pages}{436--449}.
\newblock


\bibitem[Wright and Uddin(2012)]%
        {wright2012organic}
\bibfield{author}{\bibinfo{person}{Matthew Wright} {and} \bibinfo{person}{Ashraf Uddin}.} \bibinfo{year}{2012}\natexlab{}.
\newblock \showarticletitle{Organic—inorganic hybrid solar cells: A comparative review}.
\newblock \bibinfo{journal}{\emph{Solar energy materials and solar cells}}  \bibinfo{volume}{107} (\bibinfo{year}{2012}), \bibinfo{pages}{87--111}.
\newblock


\bibitem[Wu et~al\mbox{.}(2024)]%
        {wu2024netllm}
\bibfield{author}{\bibinfo{person}{Duo Wu}, \bibinfo{person}{Xianda Wang}, \bibinfo{person}{Yaqi Qiao}, \bibinfo{person}{Zhi Wang}, \bibinfo{person}{Junchen Jiang}, \bibinfo{person}{Shuguang Cui}, {and} \bibinfo{person}{Fangxin Wang}.} \bibinfo{year}{2024}\natexlab{}.
\newblock \showarticletitle{NetLLM: Adapting Large Language Models for Networking}. In \bibinfo{booktitle}{\emph{Proceedings of the ACM SIGCOMM 2024 Conference}}. \bibinfo{pages}{661--678}.
\newblock


\bibitem[Wu et~al\mbox{.}(2020)]%
        {wu2020comprehensive}
\bibfield{author}{\bibinfo{person}{Zonghan Wu}, \bibinfo{person}{Shirui Pan}, \bibinfo{person}{Fengwen Chen}, \bibinfo{person}{Guodong Long}, \bibinfo{person}{Chengqi Zhang}, {and} \bibinfo{person}{S~Yu Philip}.} \bibinfo{year}{2020}\natexlab{}.
\newblock \showarticletitle{A comprehensive survey on graph neural networks}.
\newblock \bibinfo{journal}{\emph{IEEE transactions on neural networks and learning systems}} \bibinfo{volume}{32}, \bibinfo{number}{1} (\bibinfo{year}{2020}), \bibinfo{pages}{4--24}.
\newblock


\bibitem[Xu et~al\mbox{.}(2024)]%
        {xu2024mental}
\bibfield{author}{\bibinfo{person}{Xuhai Xu}, \bibinfo{person}{Bingsheng Yao}, \bibinfo{person}{Yuanzhe Dong}, \bibinfo{person}{Saadia Gabriel}, \bibinfo{person}{Hong Yu}, \bibinfo{person}{James Hendler}, \bibinfo{person}{Marzyeh Ghassemi}, \bibinfo{person}{Anind~K Dey}, {and} \bibinfo{person}{Dakuo Wang}.} \bibinfo{year}{2024}\natexlab{}.
\newblock \showarticletitle{Mental-llm: Leveraging large language models for mental health prediction via online text data}.
\newblock \bibinfo{journal}{\emph{Proceedings of the ACM on Interactive, Mobile, Wearable and Ubiquitous Technologies}} \bibinfo{volume}{8}, \bibinfo{number}{1} (\bibinfo{year}{2024}), \bibinfo{pages}{1--32}.
\newblock


\bibitem[Yang et~al\mbox{.}(2024)]%
        {yang2024you}
\bibfield{author}{\bibinfo{person}{Huanqi Yang}, \bibinfo{person}{Sijie Ji}, \bibinfo{person}{Rucheng Wu}, {and} \bibinfo{person}{Weitao Xu}.} \bibinfo{year}{2024}\natexlab{}.
\newblock \showarticletitle{Are You Being Tracked? Discover the Power of Zero-Shot Trajectory Tracing with LLMs!}
\newblock \bibinfo{journal}{\emph{arXiv preprint arXiv:2403.06201}} (\bibinfo{year}{2024}).
\newblock


\bibitem[Yohanna et~al\mbox{.}(2011)]%
        {yohanna2011model}
\bibfield{author}{\bibinfo{person}{Jonathan~K Yohanna}, \bibinfo{person}{Isaac~N Itodo}, {and} \bibinfo{person}{Victor~I Umogbai}.} \bibinfo{year}{2011}\natexlab{}.
\newblock \showarticletitle{A model for determining the global solar radiation for Makurdi, Nigeria}.
\newblock \bibinfo{journal}{\emph{Renewable Energy}} \bibinfo{volume}{36}, \bibinfo{number}{7} (\bibinfo{year}{2011}), \bibinfo{pages}{1989--1992}.
\newblock


\bibitem[Zhang et~al\mbox{.}(2018)]%
        {zhang2018deep}
\bibfield{author}{\bibinfo{person}{Jinsong Zhang}, \bibinfo{person}{Rodrigo Verschae}, \bibinfo{person}{Shohei Nobuhara}, {and} \bibinfo{person}{Jean-Fran{\c{c}}ois Lalonde}.} \bibinfo{year}{2018}\natexlab{}.
\newblock \showarticletitle{Deep photovoltaic nowcasting}.
\newblock \bibinfo{journal}{\emph{Solar Energy}}  \bibinfo{volume}{176} (\bibinfo{year}{2018}), \bibinfo{pages}{267--276}.
\newblock


\bibitem[Zhong et~al\mbox{.}(2020)]%
        {zhong2020random}
\bibfield{author}{\bibinfo{person}{Zhun Zhong}, \bibinfo{person}{Liang Zheng}, \bibinfo{person}{Guoliang Kang}, \bibinfo{person}{Shaozi Li}, {and} \bibinfo{person}{Yi Yang}.} \bibinfo{year}{2020}\natexlab{}.
\newblock \showarticletitle{Random erasing data augmentation}. In \bibinfo{booktitle}{\emph{Proceedings of the AAAI conference on artificial intelligence}}, Vol.~\bibinfo{volume}{34}. \bibinfo{pages}{13001--13008}.
\newblock


\bibitem[Zhu et~al\mbox{.}(2017)]%
        {zhu2017deep}
\bibfield{author}{\bibinfo{person}{Xiao~Xiang Zhu}, \bibinfo{person}{Devis Tuia}, \bibinfo{person}{Lichao Mou}, \bibinfo{person}{Gui-Song Xia}, \bibinfo{person}{Liangpei Zhang}, \bibinfo{person}{Feng Xu}, {and} \bibinfo{person}{Friedrich Fraundorfer}.} \bibinfo{year}{2017}\natexlab{}.
\newblock \showarticletitle{Deep learning in remote sensing: A comprehensive review and list of resources}.
\newblock \bibinfo{journal}{\emph{IEEE geoscience and remote sensing magazine}} \bibinfo{volume}{5}, \bibinfo{number}{4} (\bibinfo{year}{2017}), \bibinfo{pages}{8--36}.
\newblock


\bibitem[Zhuang et~al\mbox{.}(2020)]%
        {zhuang2020comprehensive}
\bibfield{author}{\bibinfo{person}{Fuzhen Zhuang}, \bibinfo{person}{Zhiyuan Qi}, \bibinfo{person}{Keyu Duan}, \bibinfo{person}{Dongbo Xi}, \bibinfo{person}{Yongchun Zhu}, \bibinfo{person}{Hengshu Zhu}, \bibinfo{person}{Hui Xiong}, {and} \bibinfo{person}{Qing He}.} \bibinfo{year}{2020}\natexlab{}.
\newblock \showarticletitle{A comprehensive survey on transfer learning}.
\newblock \bibinfo{journal}{\emph{Proc. IEEE}} \bibinfo{volume}{109}, \bibinfo{number}{1} (\bibinfo{year}{2020}), \bibinfo{pages}{43--76}.
\newblock


\end{thebibliography}
\end{document}